\documentclass{article}

\PassOptionsToPackage{numbers, compress}{natbib}

 \usepackage[main, final]{neurips_2026}

\usepackage[utf8]{inputenc} 
\usepackage[T1]{fontenc}    
\usepackage{hyperref}       
\usepackage{url}            
\usepackage{booktabs}       
\usepackage{amsfonts}       
\usepackage{nicefrac}       
\usepackage{microtype}      
\usepackage{xcolor}         

\usepackage{kotex}
\usepackage{multirow}
\usepackage{amsmath}
\usepackage{graphicx}
\usepackage{caption}
\usepackage{subcaption}
\usepackage{wrapfig}
\usepackage[table]{xcolor}
\usepackage{float}
\usepackage{pifont}
\newcommand{\cmark}{\ding{51}}
\newcommand{\xmark}{\ding{55}}

\title{Hardware-Friendly Token-Group Activation Quantization for Low-Bit Mamba Super-Resolution}

\author{%
  Jinwoo Chung \\
  Kookmin University\\
  Seoul, Republic of Korea \\
  \texttt{imaboybut@kookmin.ac.kr} \\
  \And
  Jangho Kim\thanks{Corresponding author.} \\
  Kookmin University\\
  Seoul, Republic of Korea \\
  \texttt{jangho.kim@kookmin.ac.kr} \\
}

\begin{document}

\maketitle

\newcommand\jh[1]{\textcolor{red}{#1}}
\newcommand\jw[1]{\textcolor{blue}{#1}}

\begin{abstract}
Mamba-based super-resolution models are efficient alternatives to Transformer-based restoration backbones, but their low-bit post-training quantization remains challenging because selective scan, Mamba's input-dependent state-space operator, induces different activation ranges across spatial tokens. Thus, a single layer-wise clipping range poorly matches all tokens. We propose \emph{Token-Group Quantization}, a hardware-friendly activation PTQ framework that statically approximates a hardware-unfriendly post-scan grouping oracle, where tokens with similar preferred clipping ranges are assigned to the same group. To avoid runtime post-scan grouping in static low-bit inference, a lightweight predictor infers the groups from pre-scan features available before the quantized scan path. During calibration, we form pseudo-labels by clustering post-scan activation descriptors that capture similar rounding--clipping behavior. At inference, the frozen predictor maps pre-scan features to groups and selects entries from offline-calibrated group-wise clipping-bound tables, without runtime statistics collection, dynamic clipping updates, or post-scan group assignment. To address low-bit restoration degradation, we further introduce a Frequency-Preserving Refinement (FPR) objective for clipping-bound calibration. FPR aligns Fourier amplitude and phase between full-precision teacher and quantized student outputs to preserve high-frequency structures. Across PTQ protocols, bit-widths, backbones, and tasks, our method consistently outperforms prior Mamba and SR PTQ baselines, transfers to Transformer SR and Vision Mamba classification, composes with rotation-based quantization, and achieves $2.23\times$ speedup on Jetson Orin Nano over FP16. \href{https://github.com/imaboybut/TGQ-MambaSR}{\textcolor{green!50!black}{Code is available here}}.

\end{abstract}

\section{Introduction}
\label{sec:intro}

Single image super-resolution (SR) reconstructs high-resolution images from
low-resolution inputs. Recent SR backbones have progressed from CNNs to
Transformers, where long-range dependency modeling improves reconstruction
fidelity~\citep{edsr,rcan,swinir,hat}. More recently, Mamba-based SR models
have emerged as efficient alternatives by replacing quadratic self-attention
with selective state-space modeling, capturing long-range spatial dependencies
with linear complexity~\citep{mamba,mambair,mambairv2}. Selective scan is the
core input-dependent state-space operator in this family, propagating
information across flattened spatial tokens; in MambaIRv2, our primary
backbone, it is implemented inside Attentive State-Space Modules (ASSMs).
While this improves architectural efficiency, edge deployment still requires
reducing activation memory traffic and enabling low-bit fixed-point execution,
motivating the study of post-training quantization (PTQ) for Mamba-based SR.

Although PTQ compresses pretrained models without updating the backbone
weights~\citep{adaround,brecq}, conventional PTQ severely degrades low-bit
Mamba SR. We identify a key cause. Selective scan substantially amplifies
token-wise activation heterogeneity. As shown in Sec.~\ref{sec:obs}, in MambaIRv2-Light, 18 out of 20 ASSM
blocks show increased per-token range variation after selective scan, with an
average $1.80\times$ increase over pre-scan activations. Thus, the bottleneck is not a few rare outliers, but broad token-wise range
heterogeneity: a loose range wastes precision on typical tokens, while a tight
range clips high-response tokens critical for restoration.

This mismatch suggests that different token groups should be quantized under
different clipping ranges. The most direct way to obtain such groups is to
inspect post-scan activations at the target quantization sites and assign each
token to a group accordingly, since these activations reveal the clipping range
preferred at those sites. However, this post-scan group assignment is
hardware-unfriendly for static low-bit deployment. It requires runtime statistics
collection inside the quantized path and complicates ahead-of-time compilation
with low-bit fixed-point kernels. We call this hardware-unfriendly upper-bound
grouping rule the \emph{downstream oracle}, a post-scan grouping oracle
instantiated by K-means over post-scan activation descriptors that encode
rounding--clipping regimes. The challenge is therefore to recover the oracle's
grouping behavior under a static deployment constraint, where clipping-bound
tables and quantization parameters remain fixed after calibration.

Building on this observation, we propose \textbf{Token-Group Quantization},
which distills the downstream oracle's grouping behavior into a static pre-scan
prediction problem. This approximation is feasible because selective scan generates
token-dependent recurrence parameters from pre-scan features, which therefore
already encode signals predictive of downstream activation regimes. During offline
calibration, quantization-aware pseudo-labels capture token groups that should
share clipping ranges, and a lightweight shared predictor learns to assign
tokens to these groups from pre-scan features. At inference, the frozen predictor selects entries from clipping-bound tables indexed by layer,
quantization site, and group, avoiding runtime statistics collection, dynamic clipping updates, and
post-scan group assignment.

Beyond activation range mismatch, low-bit SR quantization also degrades
high-frequency edges and textures, which pixel-space losses can
under-supervise. To address this restoration-specific degradation, we introduce
a teacher-supervised Frequency-Preserving Refinement (FPR) objective for
clipping-bound calibration, which aligns Fourier amplitude and phase between
full-precision teacher and quantized student outputs. Our main contributions are summarized as
follows:
\begin{itemize}
    \item We identify selective-scan-induced token-wise activation heterogeneity as a key failure mode of low-bit PTQ for Mamba-based SR, showing through pre- and post-scan analysis that selective scan consistently amplifies token-range variation across most ASSM blocks.

\item We propose \emph{Token-Group Quantization}, which frames token-adaptive
low-bit quantization as a static approximation to a hardware-unfriendly
post-scan grouping oracle. During calibration, we instantiate this oracle by
K-means over downstream activation descriptors and train a lightweight pre-scan
predictor to recover oracle-aligned groups before the quantized sites.

    \item We introduce a teacher-supervised Frequency-Preserving Refinement
    (FPR) objective for low-bit SR PTQ, which refines clipping bounds by
    aligning Fourier amplitude and phase between full-precision teacher and
    quantized student outputs. Experiments show consistent gains over prior
    Mamba and SR PTQ baselines with $2.23\times$ speedup on Jetson Orin Nano.
    
    \item The gains hold across MambaIR v1 and v2 at Light and Base scales and across denoising and JPEG artifact removal, transfer to SwinIR-light and to ImageNet on Vision Mamba, and compose with channel-axis transforms such as SpinQuant, as shown in Appendices~\ref{app:additional_results} and~\ref{app:beyond_mamba_sr}.
\end{itemize}

\section{Related work}

\subsection{Image Super-Resolution}
Single image super-resolution recovers high-resolution images from low-resolution inputs.
Early SR models were dominated by CNNs such as SRCNN~\cite{srcnn}, VDSR~\cite{vdsr}, EDSR~\cite{edsr}, and RCAN~\cite{rcan}, while Transformer-based models such as SwinIR~\cite{swinir} and HAT~\cite{hat} improve restoration quality by modeling long-range dependencies at higher compute and memory cost.
State-space models with selective scan~\cite{mamba} have been introduced to vision~\citep{visionmamba,vmamba} as efficient long-range backbones, and MambaIR~\cite{mambair} and MambaIRv2~\cite{mambairv2} adapt them to image restoration by combining attention, state-space blocks, and convolution.
Despite their efficiency, edge deployment of Mamba-based SR remains challenging due to activation memory traffic and low-bit quantization sensitivity.

\subsection{Post-Training Quantization}
Neural network quantization reduces deployment cost and is commonly divided into quantization-aware training (QAT) and post-training quantization (PTQ).
QAT simulates quantization during training and can achieve high accuracy, but requires costly retraining and careful tuning~\cite{lsq, ewgs, dorefa, pact}, including recent SSM restoration work such as Q-MambaIR~\cite{qmambair}.
PTQ instead keeps the backbone weights fixed and calibrates quantization
parameters from a small dataset~\cite{aciq,adaround,brecq,minmax}, but low-bit
PTQ is vulnerable to activation outliers and input-dependent distribution shifts.
Recent studies further go beyond simulated (fake) quantization and
demonstrate actual low-bit acceleration on hardware~\cite{zhao2024atom,ashkboos2024quarot,chung2026quantized,chung2026taq}.

For SR, where activation statistics vary across spatial locations and tokens, prior methods use clipping-bound calibration and outlier-aware designs~\cite{ptq4sr,2DQuantsr,outlieraware}.

Mamba-specific PTQ methods such as PTQ4VM~\cite{ptq4vm}, Quamba~\cite{quamba}, and MambaQuant~\cite{mambaquant} target selective state-space architectures, but are mainly validated outside SR restoration and still degrade substantially on Mamba-based SR.
Weight-training PTQ methods such as AIQViT~\cite{lora_vit} and SPR$^2$Q~\cite{spr} compensate quantization error with LoRA-style adapters~\cite{lora} fused before deployment.
Rotation, affine, and scaling transforms from LLM and diffusion quantization~\citep{ashkboos2024quarot,liu2025spinquant,ma2024affinequant,sun2024flatquant,xiao2023smoothquant,li2024svdquant} remove channel-wise outliers, and Quamba and MambaQuant apply the same idea to Mamba.
All of these assign the quantization range independently of token identity, which is sound when token distributions are homogeneous but is exactly the assumption that selective scan breaks in Sec.~\ref{sec:obs}. Atom~\citep{zhao2024atom} acts on the token axis but relies on runtime statistics.
Token-Group Quantization instead assigns ranges per token group statically through offline-calibrated clipping-bound tables and a lightweight pre-scan predictor, and it is complementary to channel-axis transforms, as combining it with rotation-based methods such as SpinQuant yields further gains in Appendix~\ref{app:transforms}.

\section{Preliminaries}
\label{sec:prelim}

\subsection{Selective Scan in Mamba-based SR}

A discretized state-space model updates a hidden state $h_t$ and produces an
output $y_t$ as
\begin{equation}
h_t = \bar{A}_t h_{t-1} + \bar{B}_t x_t, \qquad
y_t = C_t h_t + D x_t .
\label{eq:ssm}
\end{equation}
Unlike linear time-invariant SSMs, Mamba makes recurrence input-dependent by
generating token-specific parameters from $x_t$. Thus, different tokens can
induce different recurrence dynamics.

In Mamba-based SR, selective scan is applied to flattened spatial tokens inside
restoration blocks, where an image feature map is represented as a sequence of
length $L=HW$. Since the recurrence parameters are token-dependent, selective
scan can produce different post-scan activation regimes across spatial
locations. This property motivates our analysis of token-wise post-scan
activation heterogeneity in Section~\ref{sec:obs}.

\subsection{Uniform Quantization for PTQ}

Post-training quantization (PTQ) converts a pretrained full-precision model into
a low-bit model by calibrating quantization parameters while keeping the
backbone weights fixed. For an activation tensor $x$, we use a $b$-bit uniform
quantizer with lower and upper clipping bounds $l$ and $u$:
\begin{equation}
Q(x; l, u, b)
=
s \cdot
\mathrm{clip}\left(
\left\lfloor \frac{x-l}{s} \right\rceil,
0, 2^b-1
\right)
+ l,
\qquad
s = \frac{u-l}{2^b-1}.
\label{eq:quantizer}
\end{equation}

In conventional layer-wise activation PTQ, a single set of clipping bounds is
shared by all tokens at each layer and quantization site:
\begin{equation}
\hat{x}^{(\ell,m)}_i
=
Q\left(x^{(\ell,m)}_i; l_{\ell,m}, u_{\ell,m}, b\right),
\label{eq:layerwise_quant}
\end{equation}
where $\ell$ denotes the layer, $m$ denotes the activation quantization site,
and $i$ denotes the spatial token. The bounds
$(l_{\ell,m},u_{\ell,m})$ are calibrated offline and fixed during inference.
Therefore, layer-wise activation PTQ is effective only when one set of bounds
can represent the activation distributions of all tokens at the same
quantization site.

\section{Observations}
\label{sec:obs}

This section analyzes how selective scan affects token-wise activation
distributions in Mamba-based restoration backbones. We study 20 ASSM blocks of
MambaIRv2-Light and collect pre-scan and post-scan activations from FP32 ASSM
modules using low-resolution DIV2K~\cite{DIV2K} validation patches.

\begin{figure*}[t]
  \centering
  \vspace{-6pt}

  \begin{subfigure}[t]{0.4\textwidth}
    \centering
    \includegraphics[width=\linewidth]{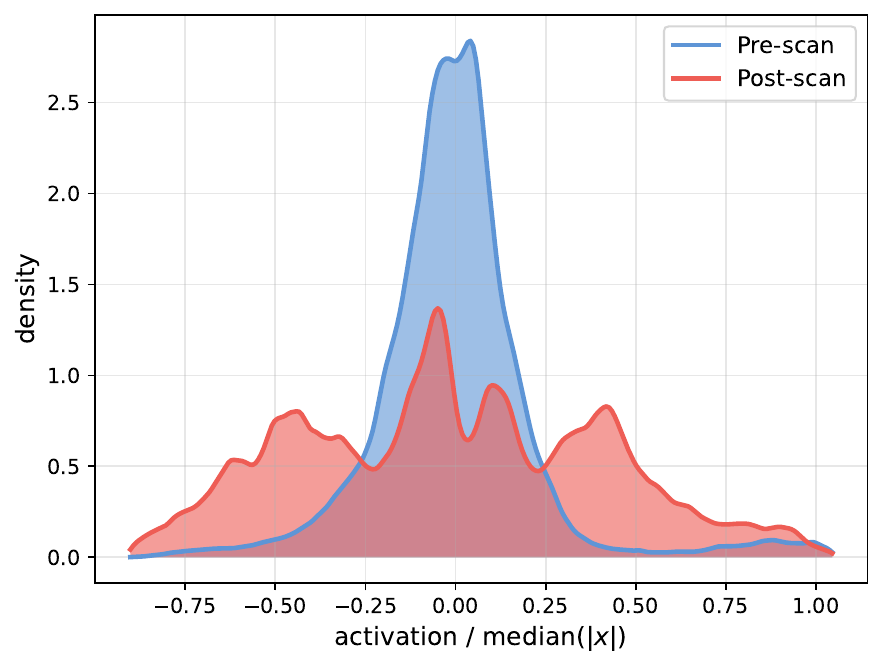}
    \caption{Activation distribution.}
    \label{fig:heterogeneity_density}
  \end{subfigure}
  \hfill
  \begin{subfigure}[t]{0.57\textwidth}
    \centering
    \includegraphics[width=\linewidth]{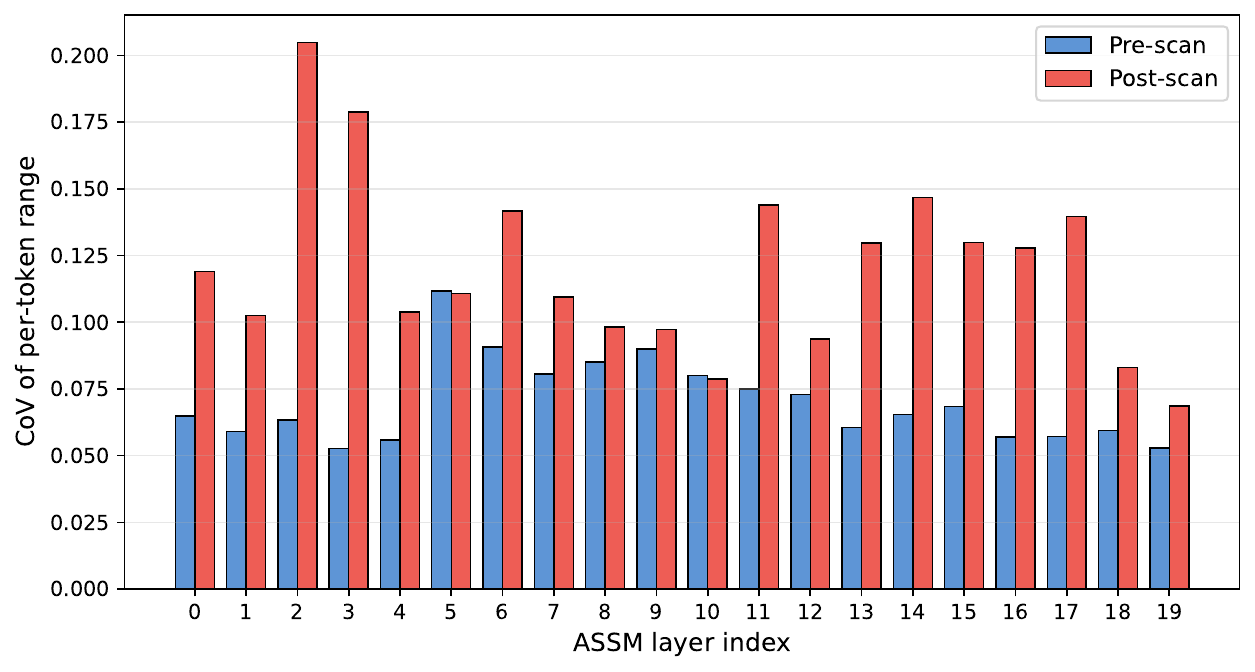}
    \caption{Layer-wise token-range CoV.}
    \label{fig:heterogeneity_cov}
  \end{subfigure}

  \vspace{-4pt}
\caption{
\textbf{Selective scan induces token-wise activation heterogeneity.}
(a) Post-scan activations exhibit a broader distribution than pre-scan activations
in a representative ASSM layer (layer 18).
(b) Per-token range CoV increases after selective scan in most ASSM layers.
}
  \label{fig:token_heterogeneity}
  \vspace{-8pt}
\end{figure*}

\subsection{Selective Scan Induces Token-wise Activation Heterogeneity}
\label{sec:obs:o1}

The central question for low-bit activation quantization is whether a single
set of layer-wise clipping bounds can represent all tokens at the same
quantization site. In Mamba-based restoration backbones, selective scan
undermines this assumption by producing token-dependent activation regimes.

\paragraph{Distributional expansion.}
Fig.~\ref{fig:token_heterogeneity}(a) shows that post-scan activations spread
over a broader range than pre-scan activations after median-$|x|$ normalization.
This indicates that selective scan disperses tokens into different magnitude
regimes, making one clipping range poorly matched to all tokens.

\paragraph{Per-token range variation.}
For each token $i$, we define its activation range as
$r_i=\max_j x_{i,j}-\min_j x_{i,j}$ and measure token-wise dispersion by
\begin{equation}
\mathrm{CoV}(r)=\frac{\mathrm{std}_i(r_i)}{\mathrm{mean}_i(r_i)}.
\end{equation}
As shown in Fig.~\ref{fig:token_heterogeneity}(b), post-scan CoV exceeds
pre-scan CoV in \textbf{18 out of 20} ASSM layers, with an average amplification
of about $1.80\times$. Thus, token-wise range variation is a consistent effect
of selective scan rather than a layer-specific artifact.

\paragraph{Conflict with low-bit quantization.}
With a single layer-wise clipping bound, fitting high-response tokens enlarges
the step size and wastes resolution for typical tokens, whereas fitting typical
tokens severely clips high-response tokens. This trade-off becomes more severe
as the bit-width decreases. The issue is therefore not isolated outliers, but
token populations with distinct activation ranges. This motivates considering group-wise clipping bounds for low-bit Mamba SR.

\begin{figure}[t]
  \centering
  \includegraphics[width=0.99\linewidth]{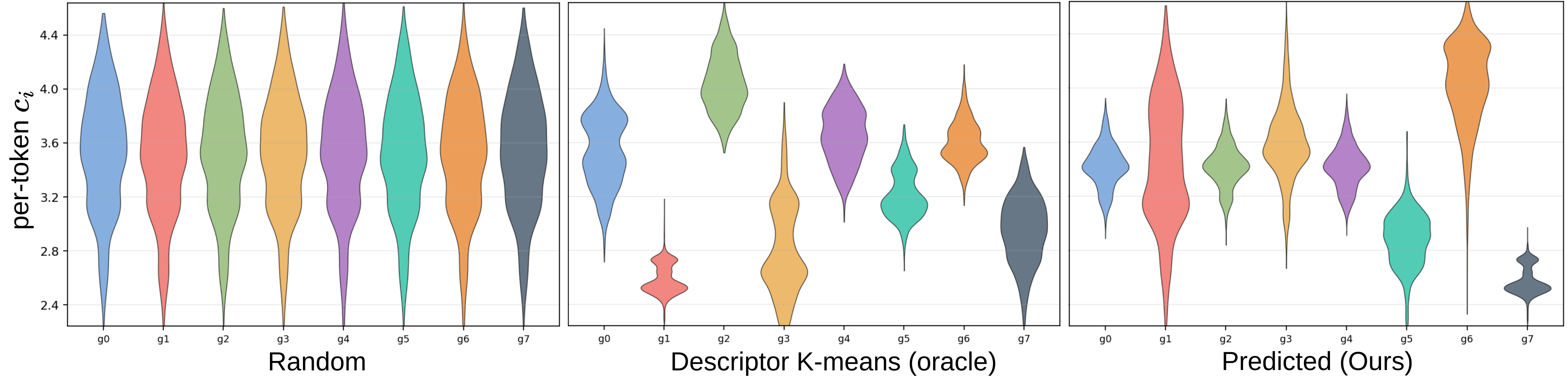}
  \caption{
  \textbf{Clipping-scale regimes.}
  Violin plots show token-level preferred clipping scales $c_i^\star$ under
  random, post-scan descriptor K-means (oracle), and predicted groups.
  }
  \label{fig:group_regime}
\end{figure}

\subsection{From Observation to Design}
\label{sec:obs:constraints}

Section~\ref{sec:obs:o1} shows that selective scan makes a single layer-wise
clipping range poorly matched to all tokens. However, larger token-wise variation
alone does not guarantee useful groups. For low-bit PTQ, groups should be defined
by \emph{quantization compatibility}: tokens should share a clipping range only
when they prefer similar downstream rounding--clipping trade-offs.

Fig.~\ref{fig:group_regime} visualizes this criterion using the token-level
preferred clipping scale
$c_i^\star=\arg\min_c\|x_i-Q(x_i;-c,c,b)\|_2^2$, a symmetric proxy for the
clipping range preferred by token $i$. Random grouping yields similar
clipping-scale distributions across groups, whereas post-scan descriptor
K-means separates tokens into distinct clipping-scale regimes. The pre-scan predicted groups recover part of this hardware-unfriendly oracle structure without observing downstream activations.

The most direct realization is a \emph{downstream oracle} that assigns groups
from post-scan activations, but this is hardware-unfriendly for static low-bit
deployment because it requires runtime statistics inside the quantized path. A
static approximation must therefore satisfy two constraints: \textbf{C1}, group
indices must be produced by a frozen function without runtime statistics; and
\textbf{C2}, group indices must be predicted before the target grouped
quantization sites while being supervised by labels derived from their
post-scan activation behavior.

These constraints rule out simpler alternatives. Layer-wise PTQ ignores token
heterogeneity; runtime post-scan grouping violates C1; per-token dynamic clipping
requires irregular scale metadata; and pre-scan clustering is deployable but not
aligned with the downstream rounding--clipping trade-off, violating C2. Thus, we
use pre-scan group prediction, downstream-aware targets, and fixed clipping-bound
tables. This is feasible because pre-scan features generate token-dependent
selective-scan parameters and therefore provide the earliest deployable signal
for approximating the oracle.

\section{Method: Token-Group Quantization}
\label{sec:method}

\begin{figure*}[t]  
  \centering
  \includegraphics[width=\textwidth]{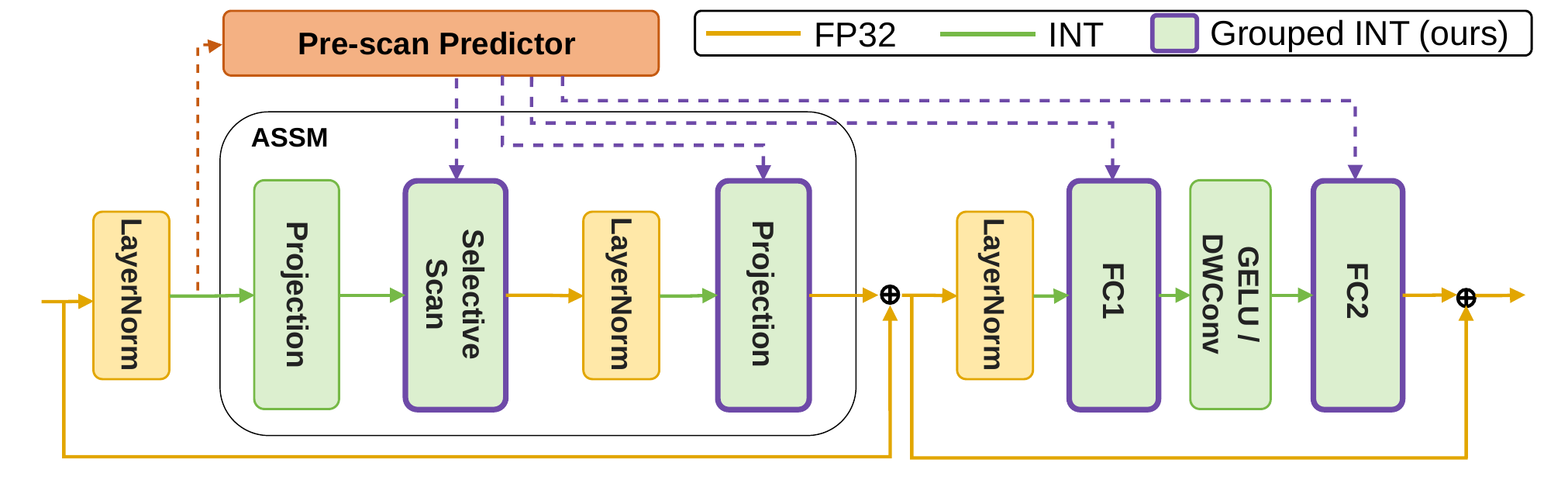} 
\caption{
Overview of Token-Group Quantization. A frozen pre-scan predictor predicts token
groups, and grouped INT sites use offline-calibrated clipping-bound tables
without runtime statistics.
}
\label{fig:overall}
\end{figure*}

Section~\ref{sec:obs:constraints} motivates a static approximation to the post-scan grouping oracle. Token-Group Quantization implements this design with three
components: a group-conditional activation quantizer, downstream-aware
pseudo-labels, and a lightweight pre-scan predictor. At inference, the frozen
predictor selects entries from fixed clipping-bound tables without runtime
statistics or post-scan group assignment. Fig.~\ref{fig:overall} illustrates the
pipeline.

\subsection{Group-Conditional Activation Quantizer}
\label{sec:method:scope}
We apply group-wise clipping bounds to the activation sites where
token-dependent distributions are most pronounced. In each block, this includes
eight group-conditional sites: three downstream linear sites and five
scan-internal per-token operands. The remaining quantized sites use
layer-wise PTQ.

Given the predicted group assignment $\hat{g}_i^{(\ell)}$, token $i$ at layer
$\ell$ and quantization site $m$ is quantized as
\begin{equation}
\hat{x}_i^{(\ell,m)}
=
Q\left(
x_i^{(\ell,m)};
l_{\ell,m,\hat{g}_i^{(\ell)}},
u_{\ell,m,\hat{g}_i^{(\ell)}},
b
\right).
\label{eq:groupquant}
\end{equation}
The bounds $(l_{\ell,m,g},u_{\ell,m,g})$ form a layer-, site-, and
group-specific clipping-bound table.

\subsection{Quantization-Aware Pseudo-Labels}
\label{sec:method:label}

A pre-scan predictor requires supervision, but the labels should encode
quantization-error regimes rather than semantics. We therefore construct
pseudo-labels offline from downstream activation descriptors, so that tokens in
the same group prefer similar clipping bounds at the quantization sites.

For each token and group-conditional site $m\in\mathcal{M}$, where
$\mathcal{M}$ denotes the eight sites above, we use a descriptor summarizing
the rounding--clipping trade-off:
\begin{equation}
\phi_m\big(x_i^{(\ell,m)}\big)
=
\Big[
q_{0.99}(|x_i^{(\ell,m)}|),\;
q_{0.999}(|x_i^{(\ell,m)}|),\;
\mathrm{RMS}(x_i^{(\ell,m)}),\;
\max |x_i^{(\ell,m)}|
\Big].
\label{eq:phi}
\end{equation}
Here, $q_p$ is the empirical $p$-quantile over channel entries. For each
token--layer pair $(i,\ell)$, we concatenate descriptors over
$m\in\mathcal{M}$. We then run a single K-means over the pooled token--layer
samples from the calibration set to obtain pseudo-labels
$g_i^{\star(\ell)}$. Descriptor construction and clustering are used only
offline. See Appendix~\ref{app:descriptor}.

\subsection{Pre-scan Group Prediction}
\label{sec:method:predictor}

Following C1 and C2, group indices must be predicted from pre-scan inputs
without runtime activation statistics. Let $\tilde{x}_i^{(\ell)}$ be the
pre-scan feature of token $i$ at layer $\ell$. Since selective scan generates
token-dependent recurrence parameters from these features, we implement the
predictor as a shared two-layer FFN:
\begin{equation}
\hat{g}_i^{(\ell)}
=
\arg\max_{k\in\{1,\ldots,K\}}
\Big[f_\theta(\tilde{x}_i^{(\ell)})\Big]_k .
\label{eq:predictor}
\end{equation}
Although $f_\theta$ is shared, $\tilde{x}_i^{(\ell)}$ is layer-dependent,
so the assignment remains layer-adaptive. We refer to this as
structure--magnitude decoupling: the shared FFN captures recurring token-structure
cues for grouping, whereas the layer- and site-specific bound tables absorb the
actual magnitude drift induced by selective scan. See
Appendix~\ref{app:predictor}.
The predictor is trained from pre-scan inputs by
\begin{equation}
\mathcal{L}_{\mathrm{pred}}
=
\mathbb{E}_{(i,\ell)}
\left[
\mathrm{CE}
\left(
f_\theta(\tilde{x}_i^{(\ell)}),
g_i^{\star(\ell)}
\right)
\right].
\label{eq:pred_loss}
\end{equation}

\subsection{Bound Initialization and Teacher Distillation}
\label{sec:method:refine}

After training and freezing the pre-scan predictor, we replay the calibration
set and collect activations according to the predicted group assignment
$\hat{g}_i^{(\ell)}$. For each $(\ell,m,g)$, we form the activation pool
assigned to group $g$:
\begin{equation}
\mathcal{X}_{\ell,m}(g)
=
\big\{
x_i^{(\ell,m)}:\hat{g}_i^{(\ell)}=g
\big\}.
\label{eq:set_group}
\end{equation}
Then we initialize the clipping bounds by minimizing empirical quantization
error:
\begin{equation}
(l_{\ell,m,g},u_{\ell,m,g})
=
\arg\min_{l,u}
\mathbb{E}_{x\sim\mathcal{X}_{\ell,m}(g)}
\left[
\left\|x-Q(x;l,u,b)\right\|_2^2
\right].
\label{eq:mse_clip}
\end{equation}

Starting from these bounds, we refine only the clipping-bound tensors using a
full-precision teacher. Given teacher/student outputs $\mathbf{y}$,
$\hat{\mathbf{y}}$ and features $F_h,\hat{F}_h$ at distillation layers
$h\in\mathcal{H}$, we use
\begin{equation}
\mathcal{L}_{\mathrm{pix}}
=
\|\hat{\mathbf{y}}-\mathbf{y}\|_1,
\qquad
\mathcal{L}_{\mathrm{feat}}
=
\frac{1}{|\mathcal{H}|}
\sum_{h\in\mathcal{H}}
\left\|
\frac{\hat{F}_h}{\|\hat{F}_h\|_2}
-
\frac{F_h}{\|F_h\|_2}
\right\|_2^2 .
\label{eq:distill_loss}
\end{equation}

\subsection{Frequency-Preserving Refinement for Bound Calibration}
\label{sec:method:fpr}

Low-bit activation quantization can attenuate textures and shift edges, while
pixel-space distillation is often dominated by low-frequency energy. Motivated
by the effectiveness of frequency-domain objectives in image
restoration~\cite{fffloss,fffloss2}, we formulate Frequency-Preserving
Refinement (FPR) as a teacher-supervised objective for SR PTQ. FPR calibrates
clipping bounds by aligning Fourier amplitude and phase between full-precision
teacher and quantized student outputs.

We apply a 2D Hann window $w$ and compute 2D Fourier transforms
$\mathbf{Y}=\mathcal{F}(\mathbf{y}\odot w)$ and
$\hat{\mathbf{Y}}=\mathcal{F}(\hat{\mathbf{y}}\odot w)$, then decompose
coefficients into amplitude $|\mathbf{Y}|$ and phase
$\angle\mathbf{Y}\in[-\pi,\pi]$:
\begin{equation}
\mathcal{L}_{\mathrm{FPR}} =
\tfrac{1}{2}\underbrace{\big\|\,|\mathbf{Y}| - |\hat{\mathbf{Y}}|\,\big\|_1}_{\text{amplitude}} +
\tfrac{1}{2}\underbrace{\big\|\,\mathrm{wrap}(\angle \mathbf{Y} - \angle \hat{\mathbf{Y}})\,\big\|_1}_{\text{phase}},
\label{eq:fpr}
\end{equation}
where $\mathrm{wrap}(\cdot)$ maps the phase difference to $(-\pi,\pi]$.
Amplitude matching preserves high-frequency energy, while phase matching
penalizes structural shifts in edges and textures.

Combining the three distillation terms gives the final objective:
\begin{equation}
\mathcal{L} = \mathcal{L}_{\mathrm{pix}} +
\lambda_{\mathrm{feat}}\, \mathcal{L}_{\mathrm{feat}} +
\lambda_{\mathrm{FPR}}\, \mathcal{L}_{\mathrm{FPR}}.
\label{eq:total_loss}
\end{equation}
Only clipping-bound tensors are optimized, while backbone weights, pseudo-labels,
predictor parameters, and bit-widths remain fixed.

\begin{table*}[!t]
    \caption{Quantitative comparison under the Weight-freeze PTQ protocol on MambaIRv2-Light.}
  \label{tab:main_table}
  \centering
  \scriptsize
  \setlength{\tabcolsep}{3pt}
  \renewcommand{\arraystretch}{0.90}

  \begin{tabular}{l l c
  c@{\hspace{4pt}}c
  c@{\hspace{4pt}}c
  c@{\hspace{4pt}}c
  c@{\hspace{4pt}}c
  c@{\hspace{4pt}}c}
    \toprule
     \multirow{2}{*}{Scale} & \multirow{2}{*}{Method} & \multirow{2}{*}{W / A}
      & \multicolumn{2}{c}{Set5}
      & \multicolumn{2}{c}{Set14}
      & \multicolumn{2}{c}{BSD100}
      & \multicolumn{2}{c}{Urban100}
      & \multicolumn{2}{c}{Manga109} \\
    \cmidrule(lr){4-5} \cmidrule(lr){6-7} \cmidrule(lr){8-9} \cmidrule(lr){10-11} \cmidrule(lr){12-13}
     & & & PSNR$\uparrow$ & SSIM$\uparrow$ & PSNR$\uparrow$ & SSIM$\uparrow$
         & PSNR$\uparrow$ & SSIM$\uparrow$ & PSNR$\uparrow$ & SSIM$\uparrow$
         & PSNR$\uparrow$ & SSIM$\uparrow$ \\
    \midrule

    \multirow{13}{*}{$\times$4}
      & MambaIRv2-Light & 32 / 32& 32.51 & 0.8992 & 28.84 & 0.7878 & 27.75 & 0.7426 & 26.82 & 0.8079 & 31.24 & 0.9182 \\
    \cmidrule(lr){2-13}
      & PTQ4VM~\cite{ptq4vm} {\textcolor{black!60}{(WACV'25)}}     & 4 / 4 & 30.82 & 0.8670 & 27.69 & 0.7546 & 26.95 & 0.7115 & 24.76 & 0.7321 & 28.19 & 0.8660 \\
      & Quamba~\cite{quamba} {\textcolor{black!60}{(ICLR’25)}}      &  4 / 4 & 31.01 & 0.8715 & 27.77 & 0.7585 & 26.99 & 0.7149 & 25.01 & 0.7470 & 28.57 & 0.8752 \\
      & MambaQuant~\cite{mambaquant} {\textcolor{black!60}{(ICLR’25)}} &  4 / 4 & 30.74 & 0.8650 & 27.17 & 0.7413 & 26.37 & 0.6920 & 23.28 & 0.6694 & 26.73 & 0.8186 \\
      & OAQ-SR~\cite{outlieraware} {\textcolor{black!60}{(ICCV’25)}}     &  4 / 4 & 30.40 & 0.8577 & 27.41 & 0.7470 & 26.75 & 0.7045 & 24.31 & 0.7136 & 27.39 & 0.8485 \\
      & \textbf{Ours} & \textbf{4 / 4} & \textbf{31.46} & \textbf{0.8811} & \textbf{28.14} & \textbf{0.7683} & \textbf{27.24} & \textbf{0.7231} & \textbf{25.39} & \textbf{0.7600} & \textbf{29.21} & \textbf{0.8884} \\
    \cmidrule(lr){2-13}
      & PTQ4VM~\cite{ptq4vm} {\textcolor{black!60}{(WACV'25)}}      & 2 / 2 & 28.77 & 0.8162 & 26.36 & 0.7167 & 26.16 & 0.6802 & 23.37 & 0.6704 & 25.26 & 0.7943 \\
      & Quamba~\cite{quamba} {\textcolor{black!60}{(ICLR’25)}}     & 2 / 2 & 28.88 & 0.8080 & 26.45 & 0.7131 & 26.20 & 0.6752 & 23.48 & 0.6651 & 25.43 & 0.7818 \\
      & MambaQuant~\cite{mambaquant} {\textcolor{black!60}{(ICLR’25)}}  & 2 / 2 & 28.84 & 0.8079 & 26.41 & 0.7114 & 26.18 & 0.6739 & 23.45 & 0.6648 & 25.38 & 0.7829 \\
      & OAQ-SR~\cite{outlieraware} {\textcolor{black!60}{(ICCV’25)}}     & 2 / 2 & 28.78 & 0.8122 & 26.37 & 0.7143 & 26.17 & 0.6780 & 23.40 & 0.6681 & 25.33 & 0.7892 \\
      & \textbf{Ours} & \textbf{2 / 2} & \textbf{29.12} & \textbf{0.8260} & \textbf{26.57} & \textbf{0.7259} & \textbf{26.32} & \textbf{0.6891} & \textbf{23.57} & \textbf{0.6805} & \textbf{25.58} & \textbf{0.8031} \\

    \midrule

    \multirow{13}{*}{$\times$2}
      & MambaIRv2-Light & 32 / 32 & 38.26 & 0.9615 & 34.09 & 0.9221 & 32.36 & 0.9019 & 33.26 & 0.9378 & 39.35 & 0.9785 \\
    \cmidrule(lr){2-13}
      & PTQ4VM~\cite{ptq4vm} {\textcolor{black!60}{(WACV'25)}}     & 4 / 4 & 37.17 & 0.9549 & 32.86 & 0.9099 & 31.57 & 0.8900 & 30.47 & 0.9084 & 37.22 & 0.9706 \\
      & Quamba~\cite{quamba} {\textcolor{black!60}{(ICLR’25)}}     &  4 / 4 & 37.07 & 0.9544 & 32.77 & 0.9092 & 31.47 & 0.8896 & 30.54 & 0.9107 & 36.94 & 0.9699 \\
      & MambaQuant~\cite{mambaquant} {\textcolor{black!60}{(ICLR’25)}}  &  4 / 4& 36.67 & 0.9495 & 31.76 & 0.8899 & 30.85 & 0.8756 & 28.08 & 0.8407 & 33.47 & 0.9186 \\
      & OAQ-SR~\cite{outlieraware} {\textcolor{black!60}{(ICCV’25)}}     &  4 / 4 & 37.22 & 0.9551 & 32.60 & 0.9071 & 31.51 & 0.8897 & 30.49 & 0.9122 & 36.91 & 0.9717 \\
      & \textbf{Ours} & \textbf{4 / 4} & \textbf{37.56} & \textbf{0.9578} & \textbf{33.24} & \textbf{0.9149} & \textbf{31.86} & \textbf{0.8950} & \textbf{31.34} & \textbf{0.9204} & \textbf{37.85} & \textbf{0.9742} \\
    \cmidrule(lr){2-13}
      & PTQ4VM~\cite{ptq4vm} {\textcolor{black!60}{(WACV'25)}}     & 2 / 2  & 34.38 & 0.9328 & 31.05 & 0.8886 & 30.21 & 0.8660 & 27.61 & 0.8603 & 32.04 & 0.9399 \\
      & Quamba~\cite{quamba} {\textcolor{black!60}{(ICLR’25)}}     &2 / 2 & 34.66 & 0.9339 & 31.26 & 0.8899 & 30.38 & 0.8687 & 27.80 & 0.8613 & 32.50 & 0.9407 \\
      & MambaQuant~\cite{mambaquant} {\textcolor{black!60}{(ICLR’25)}} &2 / 2& 34.65 & 0.9337 & 31.22 & 0.8885 & 30.36 & 0.8685 & 27.78 & 0.8610 & 32.43 & 0.9395 \\
      & OAQ-SR~\cite{outlieraware} {\textcolor{black!60}{(ICCV’25)}}     &2 / 2 & 34.54 & 0.9332 & 31.19 & 0.8907 & 30.33 & 0.8705 & 27.71 & 0.8613 & 32.29 & 0.9408 \\
      & \textbf{Ours} & \textbf{2 / 2} & \textbf{35.02} & \textbf{0.9436} & \textbf{31.41} & \textbf{0.8966} & \textbf{30.51} & \textbf{0.8766} & \textbf{27.90} & \textbf{0.8708} & \textbf{32.67} & \textbf{0.9508} \\
    \bottomrule
  \end{tabular}
\end{table*}

\section{Experiments}

\subsection{Experimental Setup}

\textbf{Datasets.}
We follow the standard SR PTQ protocol in prior work~\cite{ptq4sr,2DQuantsr}.
Calibration is performed using only low-resolution images without high-resolution
targets. We construct the calibration set from 96 low-resolution patches sampled
from the DF2K training split~\cite{DIV2K,edsr}. Evaluation is conducted on
Set5~\cite{set5}, Set14~\cite{set14}, BSD100~\cite{bsd100},
Urban100~\cite{urban100}, and Manga109~\cite{manga109}.

\textbf{Metrics and protocol.}
We report PSNR and SSIM~\cite{ssim} on the Y channel in YCbCr.
Following common SR practice, image boundaries are cropped by the upscaling
factor before evaluation.

\textbf{Backbone.}
We use MambaIRv2-Light~\cite{mambairv2} as the primary backbone. We additionally
evaluate MambaIRv1~\cite{mambair} to verify the token-group quantization
generalizes across selective-scan SR architectures.

\textbf{Implementation details.}
Calibration images are cropped and augmented following prior SR PTQ work~\cite{spr,2DQuantsr}.
We initialize clipping bounds by empirical quantization-error minimization
(Eq.~\ref{eq:mse_clip}) and refine the clipping bounds using Eq.~\ref{eq:total_loss}
with $\lambda_{\mathrm{feat}}{=}5$ and $\lambda_{\mathrm{FPR}}{=}0.01$.
Unless otherwise specified, we use $K{=}8$ token groups.
We use a shared two-layer FFN predictor detailed in Appendix~\ref{app:predictor_details}.
Bound refinement is performed for 3{,}000 iterations with a batch size of 8.
All experiments are conducted on three NVIDIA RTX A6000 GPUs. Reported numbers use seed 0, and repeating the full pipeline under five seeds gives std below 0.03~dB at W4A4 and 0.05~dB at W2A2 (Appendix~\ref{app:seeds}).

\begin{wraptable}{r}{0.3\linewidth}
  \vspace{-10pt}
  \centering
  \scriptsize
  \caption{FLOPs and W/A quantization scope.}
  \label{tab:flops_scope}
  \setlength{\tabcolsep}{3pt}
  \renewcommand{\arraystretch}{0.88}
  \begin{tabular}{lccc}
    \toprule
    Module & FLOPs & Ratio & Quant. \\
           & (G)   & (\%)  &        \\
    \midrule
    Linear + Conv & 8.49 & 85.9 & Yes \\
    Scan operands & 0.49 & 4.9  & Yes \\
    Other Conv    & 0.90 & 9.1  & No  \\
    \midrule
    Total         & 9.88 & 100.0 & -- \\
    \bottomrule
  \end{tabular}
  \vspace{-8pt}
\end{wraptable}

\textbf{Baselines.}
For the Weight-freeze PTQ protocol, we compare with PTQ4VM~\cite{ptq4vm}, Quamba~\cite{quamba},
MambaQuant~\cite{mambaquant}, and OAQ-SR~\cite{outlieraware}. PTQ4VM, Quamba, and MambaQuant
results are taken from SPR$^2$Q~\cite{spr}, while OAQ-SR and ours are evaluated by us.
For the Weight-training PTQ with LoRA protocol, we compare with AIQViT~\cite{lora_vit} and
SPR$^2$Q~\cite{spr}. Our Ours+LoRA follows the SPR$^2$Q rectifier training-and-fusion protocol, while
applying token-group activation quantization. All methods use the same W/A
low-bit scope. Table~\ref{tab:flops_scope} shows that this low-bit path covers
90.8\% of FLOPs, where Quant. for scan operands denotes activation-operand
quantization.

\begin{table*}[!t]
  \caption{Quantitative comparison under the Weight-training PTQ with LoRA protocol on MambaIRv2-Light.}  
  \label{tab:lora_ptq}
  \centering
  \scriptsize
  \setlength{\tabcolsep}{3pt}
  \renewcommand{\arraystretch}{0.90}

  \begin{tabular}{l l c
  c@{\hspace{4pt}}c
  c@{\hspace{4pt}}c
  c@{\hspace{4pt}}c
  c@{\hspace{4pt}}c
  c@{\hspace{4pt}}c}
    \toprule
     \multirow{2}{*}{Scale} & \multirow{2}{*}{Method} & \multirow{2}{*}{W / A}
      & \multicolumn{2}{c}{Set5}
      & \multicolumn{2}{c}{Set14}
      & \multicolumn{2}{c}{BSD100}
      & \multicolumn{2}{c}{Urban100}
      & \multicolumn{2}{c}{Manga109} \\
    \cmidrule(lr){4-5} \cmidrule(lr){6-7} \cmidrule(lr){8-9} \cmidrule(lr){10-11} \cmidrule(lr){12-13}
     & & & PSNR$\uparrow$ & SSIM$\uparrow$ & PSNR$\uparrow$ & SSIM$\uparrow$
         & PSNR$\uparrow$ & SSIM$\uparrow$ & PSNR$\uparrow$ & SSIM$\uparrow$
         & PSNR$\uparrow$ & SSIM$\uparrow$ \\
    \midrule

    \multirow{8}{*}{$\times$4}
      & MambaIRv2-Light & 32 / 32  & 32.51 & 0.8992 & 28.84 & 0.7878 & 27.75 & 0.7426 & 26.82 & 0.8079 & 31.24 & 0.9182 \\
    \cmidrule(lr){2-13}
      & AIQViT~\cite{lora_vit} {\textcolor{black!60}{(AAAI'25)}}             & 4 / 4 & 31.51 & 0.8856 & 28.10 & 0.7707 & 27.26 & 0.7268 & 25.18 & 0.7546 & 28.76 & 0.8854 \\
      & SPR$^2$Q~\cite{spr} {\textcolor{black!60}{(ICLR'26)}} & 4 / 4 & 32.04 & 0.8929 & 28.59 & 0.7812 & 27.52 & 0.7359 & 26.07 & 0.7864 & 30.12 & 0.9048 \\
      & \textbf{Ours + LoRA} & \textbf{4 / 4} & \textbf{32.25} & \textbf{0.8953} & \textbf{28.67} & \textbf{0.7830} & \textbf{27.61} & \textbf{0.7377} & \textbf{26.24} & \textbf{0.7906} & \textbf{30.49} & \textbf{0.9095} \\
    \cmidrule(lr){2-13}
      & AIQViT~\cite{lora_vit} {\textcolor{black!60}{(AAAI'25)}}             & 2 / 2 & 30.52 & 0.8667 &27.44 &0.7541 & 26.79 &0.7102 & 24.35 & 0.7210 &27.27 & 0.8515 \\
      & SPR$^2$Q~\cite{spr} {\textcolor{black!60}{(ICLR'26)}} &  2 / 2  & 30.43 & 0.8635 & 27.45 & 0.7531 & 26.86 & 0.7110 & 24.37 & 0.7215 & 27.30 & 0.8529 \\
      & \textbf{Ours + LoRA} & \textbf{ 2 / 2 } & \textbf{30.93} & \textbf{0.8749} & \textbf{27.80} & \textbf{0.7626} & \textbf{27.06} & \textbf{0.7190} & \textbf{24.72} & \textbf{0.7365} & \textbf{28.11} & \textbf{0.8709} \\

    \midrule

    \multirow{8}{*}{$\times$2}
      & MambaIRv2-Light & 32 / 32 & 38.26 & 0.9615 & 34.09 & 0.9221 & 32.36 & 0.9019 & 33.26 & 0.9378 & 39.35 & 0.9785 \\
    \cmidrule(lr){2-13}
      & AIQViT~\cite{lora_vit} {\textcolor{black!60}{(AAAI'25)}}  & 4 / 4 & 37.73 & 0.9598 & 33.19 & 0.9148 & 31.92 & 0.8968 & 30.97 & 0.9167 & 37.59 & 0.9750 \\
      & SPR$^2$Q~\cite{spr} {\textcolor{black!60}{(ICLR'26)}} & 4 / 4 & 37.84 & 0.9601 & 33.42 & 0.9171 & 31.99 & 0.8972 & 32.09 & 0.9286 & 38.60 & 0.9772 \\
      & \textbf{Ours + LoRA} & \textbf{4 / 4} & \textbf{38.02} & \textbf{0.9608} & \textbf{33.77} & \textbf{0.9200} & \textbf{32.24} & \textbf{0.9005} & \textbf{32.41} & \textbf{0.9314} & \textbf{38.76} & \textbf{0.9777} \\
    \cmidrule(lr){2-13}
      & AIQViT~\cite{lora_vit} {\textcolor{black!60}{(AAAI'25)}}&  2 / 2  & 36.99 & 0.9552 & 32.35 & 0.9092 & 31.42 & 0.8907 & 29.70 & 0.9007 & 36.01 & 0.9700 \\
      & SPR$^2$Q~\cite{spr} {\textcolor{black!60}{(ICLR'26)}} &  2 / 2  & 36.92 & 0.9527 & 32.38 & 0.9105 & 31.46 & 0.8910 & 30.12 & 0.9058 & 36.98 & 0.9707 \\
      & \textbf{Ours + LoRA} & \textbf{ 2 / 2 } & \textbf{37.27} & \textbf{0.9580} & \textbf{32.93} & \textbf{0.9121} & \textbf{31.72} & \textbf{0.8939} & \textbf{30.34} & \textbf{0.9090} & \textbf{37.02} & \textbf{0.9729} \\
    \bottomrule
  \end{tabular}
\end{table*}

\begin{figure*}[t]  
  \centering
  \includegraphics[width=\textwidth]{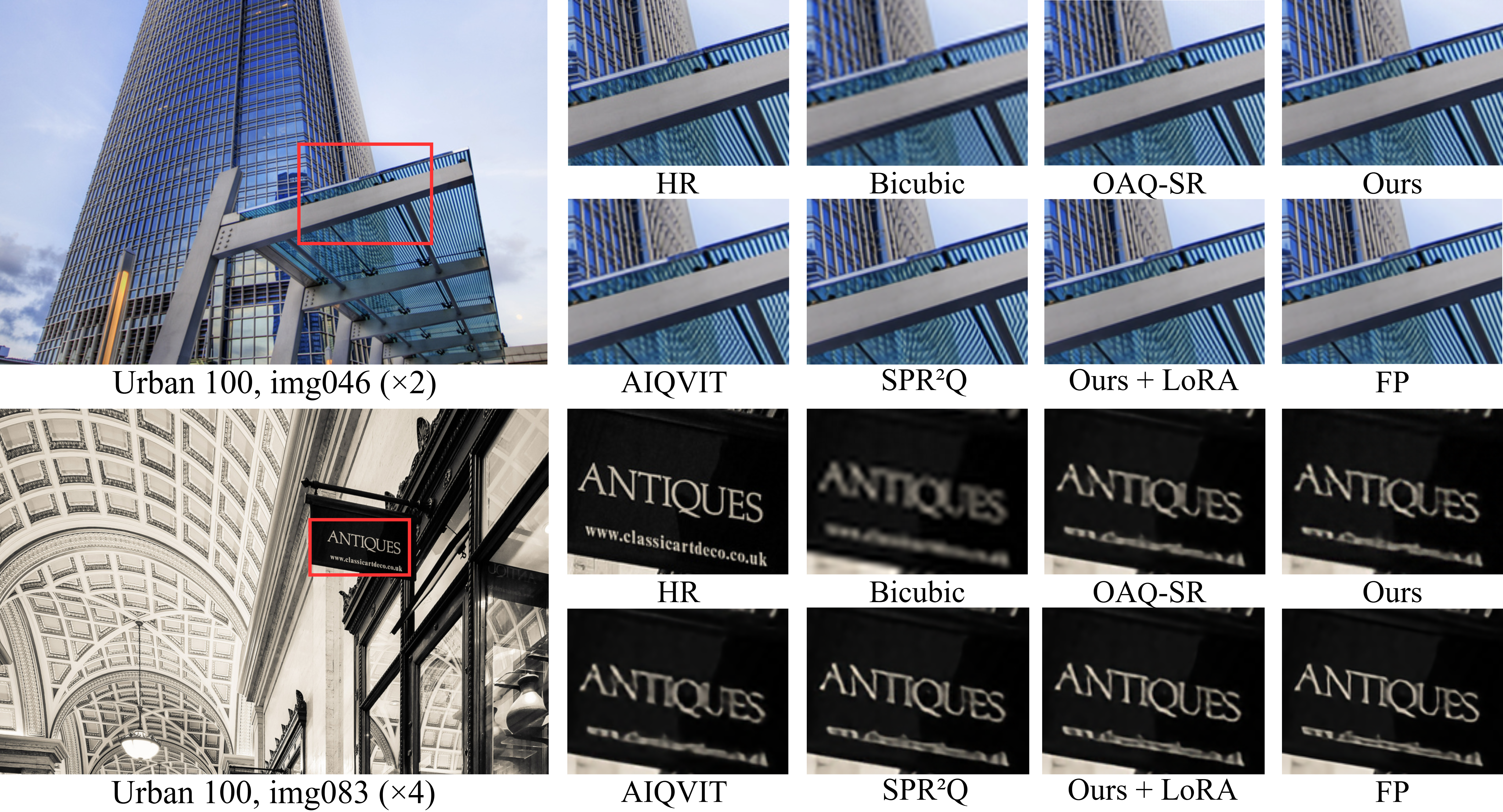} 
\caption{Visual comparison for image SR ($\times2$ and $\times4$) on Urban100 under 4-bit PTQ.}
  \label{fig_3}
\end{figure*}
\subsection{Results}
\textbf{Quantitative results.}
Table~\ref{tab:main_table} shows that our Weight-freeze PTQ consistently
outperforms Mamba-specific and SR-oriented PTQ baselines across both scales and
bit-widths. 
Table~\ref{tab:lora_ptq} evaluates the LoRA-style Weight-training protocol.
Our Ours+LoRA variant follows the rectifier training-and-fusion protocol of
SPR$^2$Q while using our token-group activation quantization. It achieves the
best results in all settings and nearly recovers full precision at 4-bit; on
BSD100, the PSNR drop is only $0.12$ dB for $\times2$ and $0.14$ dB for
$\times4$.

\textbf{Qualitative results.}
Figure~\ref{fig_3} shows that our method better preserves structural
details under low-bit quantization. In the $\times2$ example, Ours maintains the
straight canopy edges closer to the FP result, while in the $\times4$ example,
Ours reconstructs sharper text contours with less blur than the baselines.

\textbf{Offline calibration cost.}
Table~\ref{tab:calibration_cost} compares the calibration cost on
MambaIRv2-Light ($\times2$, 4-bit). In the Weight-freeze PTQ setting, our method
adds only a small overhead over OAQ-SR, increasing the wall-clock time from
1.29\,h to 1.42\,h. Under the LoRA-style Weight-training protocol, Ours+LoRA
also remains close to SPR$^2$Q, with 6.57\,h vs. 6.12\,h. These results show
that the proposed group-wise activation quantization introduces only modest
additional calibration cost.

\textbf{Inference speed.}
Table~\ref{tab:latency_cost} reports end-to-end latency of the W4A4 static
low-bit inference path with byte-aligned integer kernels and fixed
quantization codes. Our method achieves 1.70$\times$ speedup on A6000 and
2.23$\times$ on Jetson Orin Nano over FP16, nearly matching the base 4-bit
path, whereas the runtime post-scan oracle loses the speedup entirely since it
needs an extra observation pass before quantization. Appendix~\ref{app:deployment}
explains the overhead of both paths.

\begin{table*}[t]
  \centering
  \scriptsize
  \renewcommand{\arraystretch}{1.0}
  \setlength{\tabcolsep}{3pt}

  \begin{minipage}[t]{0.31\textwidth}
    \centering
    \captionof{table}{Calibration cost on MambaIRv2-Light ($\times$2, W4A4).}
          \vspace{-10pt}
    \label{tab:calibration_cost}
    \vspace{2pt}
    \begin{tabular}{lccc}
      \toprule
      \rowcolor{black!8}
      \multicolumn{4}{c}{\textbf{Weight-freeze PTQ}} \\
      Method & Wall-clock & Iter & Params \\
      \midrule
      OAQ-SR & 1.29\,h & 3{,}000 & 1.9\,k \\
      Ours   & 1.42\,h & 3{,}000 & 4.1\,k \\
      \midrule
      \rowcolor{black!8}
      \multicolumn{4}{c}{\textbf{Weight-training PTQ with LoRA}} \\
      \midrule
      SPR$^2$Q    & 6.12\,h & 12{,}500 & 617.2\,k \\
      Ours + LoRA & 6.57\,h & 12{,}500 & 619.4\,k \\
      \bottomrule
    \end{tabular}
  \end{minipage}
  \hfill
  \begin{minipage}[t]{0.33\textwidth}
    \centering
    \captionof{table}{Latency on Set5, MambaIRv2-Light ($\times$2, W4A4).}
          \vspace{-10pt}
    \label{tab:latency_cost}
    \vspace{2pt}
    \begin{tabular}{lcc}
      \toprule
      \rowcolor{black!8}
      \multicolumn{3}{c}{\textbf{NVIDIA RTX A6000}} \\
      Method & Latency (ms) & Speedup \\
      \midrule
      FP16       & 180.7 & 1.00$\times$  \\
      4-bit Oracle & 254.5 & 0.71$\times$  \\
      4-bit Base & 102.7 & 1.76$\times$ \\
      4-bit Ours & 106.2 & 1.70$\times$ \\
      \midrule
      \rowcolor{black!8}
      \multicolumn{3}{c}{\textbf{Jetson Orin Nano}} \\
      \midrule
      FP16       & 3149.6 &  1.00$\times$  \\
      4-bit Oracle &3213.9& 0.98$\times$  \\
      4-bit Base & 1380.1 & 2.28$\times$ \\
      4-bit Ours & 1410.7 & 2.23$\times$ \\
      \bottomrule
    \end{tabular}
  \end{minipage}
  \hfill
  \begin{minipage}[t]{0.28\textwidth}
    \centering
    \captionof{table}{Number of token groups $K$ (MambaIRv2-Light, $\times$2, W4A4).}
          \vspace{-10pt}
    \label{tab:k_ablation}
    \vspace{2pt}
    \begin{tabular}{c cc}
      \toprule
      $K$ & Set14 & Urban100 \\
          & PSNR / SSIM & PSNR / SSIM \\
      \midrule
      1 & 32.62 / 0.9073 & 30.59 / 0.9127 \\
      4 & 33.07 / 0.9117 & 31.03 / 0.9175 \\
      6 & 33.19 / 0.9142 & 31.28 / 0.9199 \\
      \textbf{8} & \textbf{33.24} / \textbf{0.9149} & \textbf{31.34} / \textbf{0.9204} \\
      \bottomrule
    \end{tabular}
  \end{minipage}
\end{table*}

\begin{table*}[t]
  \centering
  \scriptsize
  \renewcommand{\arraystretch}{0.9}
  \setlength{\tabcolsep}{3pt}

  \begin{minipage}[t]{0.55\textwidth}
    \centering
    \captionof{table}{Supervision ablation on MambaIRv2-Light at $\times$2 and W4A4.
    TG marks token-group quantization. Rows differ only in the refinement objective.}
          \vspace{-10pt}
    \label{tab:component_ablation}
    \vspace{2pt}
    \begin{tabular}{l c cc}
      \toprule
      \multirow{2}{*}{Refinement objective} & \multirow{2}{*}{TG}
        & Set14 & Urban100 \\
        & & PSNR / SSIM & PSNR / SSIM \\
      \midrule
      None (Eq.~\ref{eq:mse_clip} init.\ only)                 & \xmark & 32.15 / 0.8931 & 29.84 / 0.8915 \\
      $\mathcal{L}_{\mathrm{pix}}{+}\mathcal{L}_{\mathrm{feat}}$ (layer-wise PTQ) & \xmark & 32.45 / 0.9056 & 30.34 / 0.9107 \\
      \midrule
      None (Eq.~\ref{eq:mse_clip} init.\ only)                 & \cmark & 32.38 / 0.9015 & 30.56 / 0.9115 \\
      $\mathcal{L}_{\mathrm{pix}}$ only                         & \cmark & 32.53 / 0.9018 & 30.72 / 0.9123 \\
      $\mathcal{L}_{\mathrm{pix}}{+}$ feat.\ MSE (conventional KD) & \cmark & 33.01 / 0.9091 & 30.89 / 0.9136 \\
      $\mathcal{L}_{\mathrm{pix}}{+}\mathcal{L}_{\mathrm{feat}}$ & \cmark & 33.06 / 0.9107 & 30.99 / 0.9172 \\
      \quad $+$ amplitude term only                             & \cmark & 33.15 / 0.9122 & 31.14 / 0.9190 \\
      \quad $+$ phase term only                                 & \cmark & 33.10 / 0.9128 & 31.10 / 0.9183 \\
      \quad $+\,\mathcal{L}_{\mathrm{FPR}}$ (amp.\,+\,phase, \textbf{Ours}) & \cmark
        & \textbf{33.24} / \textbf{0.9149} & \textbf{31.34} / \textbf{0.9204} \\
      \bottomrule
    \end{tabular}
  \end{minipage}
  \hfill
  \begin{minipage}[t]{0.42\textwidth}
    \centering
    \captionof{table}{Grouping rule ablation (MambaIRv2-Light, $\times$2, W4A4).
    Static: no runtime statistics (C1). The post-scan oracle (gray) is a
    hardware-unfriendly upper bound; bold marks the best static rule.}
          \vspace{-10pt}
    \label{tab:grouping_rule_main}
    \vspace{2pt}
    \begin{tabular}{l c cc}
      \toprule
      \multirow{2}{*}{Rule} & \multirow{2}{*}{Static}
        & Set14 & Urban100 \\
        & & PSNR / SSIM & PSNR / SSIM \\
      \midrule
      Layer-wise PTQ     & \cmark & 32.45 / 0.9056 & 30.34 / 0.9107 \\
      Random groups      & \cmark & 31.76 / 0.8904 & 29.91 / 0.9070 \\
      Pre-scan $K$-means & \cmark & 32.84 / 0.9092 & 30.77 / 0.9161 \\
      \textcolor{black!50}{Post-scan oracle} & \textcolor{black!50}{\xmark}
        & \textcolor{black!50}{33.41 / 0.9176} & \textcolor{black!50}{31.57 / 0.9253} \\
      Predicted (Ours)   & \cmark & \textbf{33.24} / \textbf{0.9149} & \textbf{31.34} / \textbf{0.9204} \\
      \bottomrule
    \end{tabular}
  \end{minipage}
\end{table*}

\subsection{Ablation Study}
\textbf{Number of token groups.}
Table~\ref{tab:k_ablation} shows that performance improves as the number of
token groups increases, while the gain from $K=6$ to $K=8$ nearly saturates.
We therefore use $K=8$ as the default setting. On MambaIRv2-Base the optimum shifts mildly to $K{=}16$, but $K{=}8$ already
recovers 89\% of the range (Appendix~\ref{app:k_base}).

\textbf{Component analysis.}
Table~\ref{tab:component_ablation} separates the two sources of gain. Token grouping alone accounts for 77\% of the improvement over layer-wise PTQ on Set14 and 65\% on Urban100, and it helps even without refinement. On top of the strongest conventional objective, the amplitude and phase terms each help alone and their combination adds a further 0.35~dB on Urban100, showing that the two are complementary. Appendix~\ref{app:freq_evidence} shows that this gain comes from the high-frequency band that pixel and feature matching under-supervise.

\textbf{Grouping rule.}
Table~\ref{tab:grouping_rule_main} shows that the gain is not from having more
ranges: random groups fall below layer-wise PTQ, and our predicted groups
outperform pre-scan $K$-means and recover 82.3\% and 81.3\% of the oracle gain
on Set14 and Urban100 without runtime statistics.

\textbf{FPR objective weight.} A sensitivity analysis on $\lambda_\mathrm{FPR}$ is provided in Appendix~\ref{app:fpr_ablation}.

\textbf{Generalization.}
Appendices~\ref{app:additional_results} and~\ref{app:beyond_mamba_sr} extend the evaluation to MambaIRv1-Light, MambaIRv1-Base, MambaIRv2-Base, denoising, and JPEG artifact removal, to the Transformer backbone SwinIR-light against 2DQuant, FIMA-Q, APHQ-ViT, and SPR$^2$Q, and to ImageNet classification on Vision Mamba, where the token-range amplification of Sec.~\ref{sec:obs} is even larger. Ours is best in every setting. Ported Hadamard, SpinQuant, and Affine transforms trail ours by up to 2.12~dB, and SpinQuant combined with ours is best everywhere, so the token axis and the channel axis compose. Appendix~\ref{app:resolution} further shows robustness to input resolution up to 16 times the calibration length.

\section{Conclusion}
\label{sec:conclusion}

We presented Token-Group Quantization, a static yet token-adaptive PTQ framework
for low-bit Mamba-based super-resolution. By analyzing MambaIRv2-Light, we
showed that selective scan amplifies token-wise activation heterogeneity,
making conventional layer-wise clipping ranges poorly matched to all tokens.
Our method addresses this mismatch with a lightweight pre-scan predictor,
quantization-aware pseudo-labels, and offline-calibrated group-wise clipping
bounds. We further introduced a teacher-supervised FPR objective for frequency-preserving
clipping-bound refinement. Experiments demonstrate consistent gains over prior Mamba and SR PTQ baselines
under Weight-freeze and LoRA-based protocols, transfer to Transformer SR and
Vision Mamba, and compose with channel-axis transforms, while preserving
practical acceleration on Jetson Orin Nano.

\paragraph{Limitation and Future Work.}
While Token-Group Quantization improves low-bit Mamba SR accuracy and preserves
practical acceleration, its pre-scan prediction and group-wise bound lookup
introduce a small inference overhead relative to the base quantized path. Future
work may explore deployment on other on-device accelerators, such as NPUs,
DSPs, mobile GPUs, or vendor-specific low-bit runtimes, under the same
token-adaptive quantization framework.


\medskip

\bibliographystyle{plainnat}
\bibliography{ref}

\appendix

\section{Generalization Protocol}
\label{app:additional_results}

We evaluate generalization on MambaIRv1-Light, JPEG artifact removal, and color image denoising.
For these appendix experiments, we include two complementary baselines. \emph{Layer-wise PTQ}
uses the same Weight-freeze protocol and quantization scope as ours, but uses one clipping range per
layer/site instead of group-wise ranges; thus, it directly isolates the benefit of token grouping. OAQ-SR
is included as a recent SR-oriented outlier-aware PTQ baseline. Since the goal of these experiments is
to test transferability rather than to establish a new benchmark suite for every restoration task, this
combination provides a controlled baseline and a strong task-relevant PTQ baseline.

These experiments also test the robustness of pre-scan group prediction under degraded restoration
inputs. JPEG artifact removal contains spatially non-uniform compression artifacts, while color
denoising uses noisy inputs; thus, the predictor must infer useful token groups from corrupted
features rather than clean or single-degradation inputs. The consistent gains on these tasks indicate
that the proposed grouping rule remains effective under input degradation.

\begin{table}[H]
  \caption{Generalization results on MambaIRv1-Light under the Weight-freeze PTQ protocol.}
  \label{tab:mambairv1_sr}
  \centering
  \scriptsize
  \setlength{\tabcolsep}{3pt}
  \renewcommand{\arraystretch}{0.90}

  \begin{tabular}{l l c
  c@{\hspace{4pt}}c
  c@{\hspace{4pt}}c
  c@{\hspace{4pt}}c
  c@{\hspace{4pt}}c
  c@{\hspace{4pt}}c}
    \toprule
     \multirow{2}{*}{Scale} & \multirow{2}{*}{Method} & \multirow{2}{*}{W / A}
      & \multicolumn{2}{c}{Set5}
      & \multicolumn{2}{c}{Set14}
      & \multicolumn{2}{c}{BSD100}
      & \multicolumn{2}{c}{Urban100}
      & \multicolumn{2}{c}{Manga109} \\
    \cmidrule(lr){4-5} \cmidrule(lr){6-7} \cmidrule(lr){8-9} \cmidrule(lr){10-11} \cmidrule(lr){12-13}
     & & & PSNR$\uparrow$ & SSIM$\uparrow$ & PSNR$\uparrow$ & SSIM$\uparrow$
         & PSNR$\uparrow$ & SSIM$\uparrow$ & PSNR$\uparrow$ & SSIM$\uparrow$
         & PSNR$\uparrow$ & SSIM$\uparrow$ \\
    \midrule

    \multirow{7}{*}{$\times$4}
      & MambaIRv1-Light & 32 / 32
        & 32.42 & 0.8977 & 28.74 & 0.7847 & 27.68 & 0.7400 & 26.52 & 0.7983 & 30.94 & 0.9135 \\
    \cmidrule(lr){2-13}
      & Layer-wise PTQ & 4 / 4
        & 30.39 & 0.8257 & 27.45 & 0.7249 & 26.71 & 0.6811 & 24.99 & 0.7185 & 28.62 & 0.8372 \\
      & OAQ-SR~\cite{outlieraware} {\textcolor{black!60}{(ICCV'25)}} & 4 / 4
        & 30.85 & 0.8507 & 27.80 & 0.7446 & 26.93 & 0.7019 & 25.03 & 0.7392 & 28.59 & 0.8611 \\
      & \textbf{Ours} & \textbf{4 / 4}
        & \textbf{31.10} & \textbf{0.8627}
        & \textbf{27.91} & \textbf{0.7534}
        & \textbf{27.09} & \textbf{0.7097}
        & \textbf{25.23} & \textbf{0.7444}
        & \textbf{28.90} & \textbf{0.8682} \\
    \cmidrule(lr){2-13}
      & Layer-wise PTQ & 2 / 2
        & 23.10 & 0.4199 & 22.29 & 0.3725 & 22.25 & 0.3453 & 20.92 & 0.3736 & 21.99 & 0.4017 \\
      & OAQ-SR~\cite{outlieraware} {\textcolor{black!60}{(ICCV'25)}} & 2 / 2
        & 27.98 & 0.7367 & 25.68 & 0.6477 & 25.55 & 0.6133 & 23.07 & 0.6080 & 25.02 & 0.7135 \\
      & \textbf{Ours} & \textbf{2 / 2}
        & \textbf{28.33} & \textbf{0.7611}
        & \textbf{26.09} & \textbf{0.6713}
        & \textbf{25.89} & \textbf{0.6365}
        & \textbf{23.29} & \textbf{0.6278}
        & \textbf{25.19} & \textbf{0.7390} \\

    \midrule

    \multirow{7}{*}{$\times$2}
      & MambaIRv1-Light & 32 / 32
        & 38.13 & 0.9610 & 33.95 & 0.9208 & 32.31 & 0.9013 & 32.85 & 0.9349 & 39.20 & 0.9782 \\
    \cmidrule(lr){2-13}
      & Layer-wise PTQ & 4 / 4
        & 35.07 & 0.9178 & 31.83 & 0.8774 & 30.67 & 0.8581 & 30.05 & 0.8835 & 35.06 & 0.9325 \\
      & OAQ-SR~\cite{outlieraware} {\textcolor{black!60}{(ICCV'25)}} & 4 / 4
        & 36.55 & 0.9445 & 32.67 & 0.9013 & 31.46 & 0.8821 & 30.58 & 0.9042 & 36.53 & 0.9593 \\
      & \textbf{Ours} & \textbf{4 / 4}
        & \textbf{37.12} & \textbf{0.9527}
        & \textbf{33.06} & \textbf{0.9105}
        & \textbf{31.73} & \textbf{0.8910}
        & \textbf{31.06} & \textbf{0.9160}
        & \textbf{37.25} & \textbf{0.9686} \\
    \cmidrule(lr){2-13}
      & Layer-wise PTQ & 2 / 2
        & 22.57 & 0.4159 & 21.95 & 0.4086 & 21.80 & 0.3942 & 21.33 & 0.4402 & 22.47 & 0.4216 \\
      & OAQ-SR~\cite{outlieraware} {\textcolor{black!60}{(ICCV'25)}} & 2 / 2
        & 34.03 & 0.9303 & 30.76 & 0.8809 & 29.22 & 0.8471 & 27.01 & 0.8537 & 31.78 & 0.9354 \\
      & \textbf{Ours} & \textbf{2 / 2}
        & \textbf{34.40} & \textbf{0.9371}
        & \textbf{31.05} & \textbf{0.8896}
        & \textbf{30.20} & \textbf{0.8685}
        & \textbf{27.60} & \textbf{0.8624}
        & \textbf{32.07} & \textbf{0.9436} \\

    \bottomrule
  \end{tabular}
  \vspace{-7mm}
\end{table}

\begin{table}[H]
  \caption{Generalization results on color image denoising with noise level $\sigma=15$ under the Weight-freeze PTQ protocol.}
  \label{tab:ir_colordn}
  \centering
  \scriptsize
  \setlength{\tabcolsep}{4pt}
  \renewcommand{\arraystretch}{0.90}

  \begin{tabular}{l c
  c@{\hspace{4pt}}c
  c@{\hspace{4pt}}c
  c@{\hspace{4pt}}c
  c@{\hspace{4pt}}c}
    \toprule
     \multirow{2}{*}{Method} & \multirow{2}{*}{W / A}
      & \multicolumn{2}{c}{CBSD68}
      & \multicolumn{2}{c}{Kodak24}
      & \multicolumn{2}{c}{McMaster}
      & \multicolumn{2}{c}{Urban100} \\
    \cmidrule(lr){3-4} \cmidrule(lr){5-6} \cmidrule(lr){7-8} \cmidrule(lr){9-10}
     & & PSNR$\uparrow$ & SSIM$\uparrow$ & PSNR$\uparrow$ & SSIM$\uparrow$
       & PSNR$\uparrow$ & SSIM$\uparrow$ & PSNR$\uparrow$ & SSIM$\uparrow$ \\
    \midrule

    MambaIRv2-Base & 32 / 32
      & 34.15 & 0.9388 & 35.15 & 0.9363 & 35.07 & 0.9421 & 34.72 & 0.9484 \\
    \cmidrule(lr){1-10}
    Layer-wise PTQ & 4 / 4
      & 32.84 & 0.9161
      & 33.49 & 0.9098
      & 33.06 & 0.9169
      & 32.62 & 0.9229 \\
    OAQ-SR~\cite{outlieraware} {\textcolor{black!60}{(ICCV'25)}} & 4 / 4
      & 33.52 & 0.9290
      & 34.33 & 0.9243
      & 34.04 & 0.9305
      & 33.59 & 0.9371 \\
    \textbf{Ours} & \textbf{4 / 4}
      & \textbf{33.78} & \textbf{0.9335}
      & \textbf{34.69} & \textbf{0.9299}
      & \textbf{34.51} & \textbf{0.9356}
      & \textbf{34.04} & \textbf{0.9419} \\
    \cmidrule(lr){1-10}
    Layer-wise PTQ & 2 / 2
      & 27.20 & 0.7871
      & 27.00 & 0.7667
      & 24.31 & 0.7564
      & 25.44 & 0.7860 \\
    OAQ-SR~\cite{outlieraware} {\textcolor{black!60}{(ICCV'25)}} & 2 / 2
      & 31.63 & 0.8960
      & 32.06 & 0.8838
      & 31.16 & 0.8857
      & 30.25 & 0.8944 \\
    \textbf{Ours} & \textbf{2 / 2}
      & \textbf{32.41} & \textbf{0.9142}
      & \textbf{32.89} & \textbf{0.9060}
      & \textbf{31.94} & \textbf{0.9081}
      & \textbf{31.38} & \textbf{0.9174} \\
    \bottomrule
  \end{tabular}
  \vspace{-7mm}
\end{table}

\begin{table}[H]
  \caption{Generalization results on JPEG artifact removal under the Weight-freeze PTQ protocol.}
  \label{tab:ir_car}
  \centering
  \scriptsize
  \setlength{\tabcolsep}{5pt}
  \renewcommand{\arraystretch}{0.90}

  \begin{tabular}{l l c
  c@{\hspace{4pt}}c
  c@{\hspace{4pt}}c}
    \toprule
     \multirow{2}{*}{Quality} & \multirow{2}{*}{Method} & \multirow{2}{*}{W / A}
      & \multicolumn{2}{c}{Classic5}
      & \multicolumn{2}{c}{LIVE1} \\
    \cmidrule(lr){4-5} \cmidrule(lr){6-7}
     & & & PSNR$\uparrow$ & SSIM$\uparrow$ & PSNR$\uparrow$ & SSIM$\uparrow$ \\
    \midrule

    \multirow{7}{*}{$q=10$}
      & MambaIRv2-Base & 32 / 32
        & 30.37 & 0.8269 & 29.91 & 0.8301 \\
    \cmidrule(lr){2-7}
      & Layer-wise PTQ & 4 / 4
        & 28.95 & 0.7834 & 28.74 & 0.7928 \\
      & OAQ-SR~\cite{outlieraware} {\textcolor{black!60}{(ICCV'25)}} & 4 / 4
        & 29.59 & 0.8064 & 29.26 & 0.8125 \\
      & \textbf{Ours} & \textbf{4 / 4}
        & \textbf{29.93} & \textbf{0.8183}
        & \textbf{29.57} & \textbf{0.8230} \\
    \cmidrule(lr){2-7}
      & Layer-wise PTQ & 2 / 2
        & 27.57 & 0.7324 & 27.48 & 0.7474 \\
      & OAQ-SR~\cite{outlieraware} {\textcolor{black!60}{(ICCV'25)}} & 2 / 2
        & 28.06 & 0.7511 & 28.06 & 0.7686 \\
      & \textbf{Ours} & \textbf{2 / 2}
        & \textbf{28.30} & \textbf{0.7698}
        & \textbf{28.22} & \textbf{0.7851} \\

    \midrule

    \multirow{7}{*}{$q=30$}
      & MambaIRv2-Base & 32 / 32
        & 33.81 & 0.8970 & 33.73 & 0.9179 \\
    \cmidrule(lr){2-7}
      & Layer-wise PTQ & 4 / 4
        & 32.47 & 0.8752 & 32.37 & 0.8940 \\
      & OAQ-SR~\cite{outlieraware} {\textcolor{black!60}{(ICCV'25)}} & 4 / 4
        & 33.05 & 0.8871 & 33.05 & 0.9084 \\
      & \textbf{Ours} & \textbf{4 / 4}
        & \textbf{33.39} & \textbf{0.8927}
        & \textbf{33.34} & \textbf{0.9133} \\
    \cmidrule(lr){2-7}
      & Layer-wise PTQ & 2 / 2
        & 22.78 & 0.4427 & 22.70 & 0.4525 \\
      & OAQ-SR~\cite{outlieraware} {\textcolor{black!60}{(ICCV'25)}} & 2 / 2
        & 31.60 & 0.8617 & 31.45 & 0.8822 \\
      & \textbf{Ours} & \textbf{2 / 2}
        & \textbf{31.90} & \textbf{0.8719}
        & \textbf{31.90} & \textbf{0.8934} \\

    \midrule

    \multirow{7}{*}{$q=40$}
      &MambaIRv2-Base & 32 / 32
        & 34.64 & 0.9093 & 34.73 & 0.9323 \\
    \cmidrule(lr){2-7}
      & Layer-wise PTQ & 4 / 4
        & 33.34 & 0.8909 & 33.28 & 0.9110 \\
      & OAQ-SR~\cite{outlieraware} {\textcolor{black!60}{(ICCV'25)}} & 4 / 4
        & 34.02 & 0.9033 & 34.14 & 0.9262 \\
      & \textbf{Ours} & \textbf{4 / 4}
        & \textbf{34.31} & \textbf{0.9068}
        & \textbf{34.38} & \textbf{0.9292} \\
    \cmidrule(lr){2-7}
      & Layer-wise PTQ & 2 / 2
        & 30.79 & 0.8153 & 31.17 & 0.8413 \\
      & OAQ-SR~\cite{outlieraware} {\textcolor{black!60}{(ICCV'25)}} & 2 / 2
        & 32.83 & 0.8895 & 32.86 & 0.9110 \\
      & \textbf{Ours} & \textbf{2 / 2}
        & \textbf{33.11} & \textbf{0.8925}
        & \textbf{33.22} & \textbf{0.9147} \\
    \bottomrule
  \end{tabular}
\end{table}
Tables~\ref{tab:mambairv1_sr}--\ref{tab:ir_car} show that our method consistently outperforms
both baselines across backbones, tasks, and bit-widths. This suggests that the proposed static
token-group activation ranges address a general activation-heterogeneity issue in Mamba restoration
models, rather than overfitting to the main MambaIRv2-Light SR setting.

Notably, the same pre-scan prediction principle remains effective even when
the input features are degraded by noise or JPEG artifacts, suggesting that the
predictor does not rely on clean-image-specific cues but captures quantization
regimes that persist across restoration tasks.
For generalization experiments, we prioritize controlled isolation over exhaustive benchmarking.

\section{Generalization Beyond Mamba-Based SR}
\label{app:beyond_mamba_sr}

The main paper evaluates MambaIRv2-Light as the primary backbone, and
Appendix~\ref{app:additional_results} adds MambaIRv1-Light and MambaIRv2-Base
on further restoration tasks. This section tests whether the observation of
Sec.~\ref{sec:obs} and the proposed grouping principle extend to a Transformer
SR backbone, to Mamba outside restoration, to larger MambaIR backbones, and to
channel-axis quantization transforms from the LLM and diffusion literature.

\subsection{Transformer SR Backbone}
\label{app:swinir}

We apply the same framework to SwinIR-light~\citep{swinir} with the identical
quantization scope as 2DQuant~\citep{2DQuantsr} and SPR$^2$Q~\citep{spr},
replacing the post-attention MLP activation quantizers with our
group-conditional quantizers at $K{=}8$. Since SwinIR has no scan stage, the
predictor becomes a pre-site predictor trained from pre-block features. Group
indices are still produced before the target quantization sites without
runtime statistics, so the deployment constraint C1 is unchanged. All
distillation settings follow the main paper.
Table~\ref{tab:swinir_wf} compares against 2DQuant and the ViT PTQ methods
FIMA-Q~\citep{wu2025fima} and APHQ-ViT~\citep{wu2025aphq} under the
Weight-freeze protocol, and Table~\ref{tab:swinir_lora} compares against
SPR$^2$Q under the LoRA protocol. Ours is best in every cell.

On SwinIR-light we measured token-range CoV before and after attention and did
not observe the amplification induced by selective scan. Token ranges still
vary with image content, so tokens can be clustered into distinct range
regimes and both the descriptor $K$-means and the pre-site predictor operate as
designed. Since the inter-group range gap is smaller without scan
amplification, the clipping error recoverable by grouping is also smaller,
which is consistent with the more modest margins than on MambaIRv2-Light. The
gain magnitude thus tracks the degree of token heterogeneity across backbones.
The method is generally applicable and particularly beneficial for Mamba.

\begin{table}[H]
  \caption{$\times2$ SR on SwinIR-light under the Weight-freeze PTQ protocol.}
  \label{tab:swinir_wf}
  \centering
  \scriptsize
  \setlength{\tabcolsep}{3pt}
  \renewcommand{\arraystretch}{0.90}
  \begin{tabular}{l c
  c@{\hspace{4pt}}c
  c@{\hspace{4pt}}c
  c@{\hspace{4pt}}c
  c@{\hspace{4pt}}c
  c@{\hspace{4pt}}c}
    \toprule
     \multirow{2}{*}{Method} & \multirow{2}{*}{W / A}
      & \multicolumn{2}{c}{Set5}
      & \multicolumn{2}{c}{Set14}
      & \multicolumn{2}{c}{BSD100}
      & \multicolumn{2}{c}{Urban100}
      & \multicolumn{2}{c}{Manga109} \\
    \cmidrule(lr){3-4} \cmidrule(lr){5-6} \cmidrule(lr){7-8} \cmidrule(lr){9-10} \cmidrule(lr){11-12}
     & & PSNR$\uparrow$ & SSIM$\uparrow$ & PSNR$\uparrow$ & SSIM$\uparrow$
         & PSNR$\uparrow$ & SSIM$\uparrow$ & PSNR$\uparrow$ & SSIM$\uparrow$
         & PSNR$\uparrow$ & SSIM$\uparrow$ \\
    \midrule
    SwinIR-light & 32 / 32 & 38.14 & 0.9611 & 33.86 & 0.9206 & 32.31 & 0.9012 & 32.76 & 0.9340 & 39.12 & 0.9783 \\
    \midrule
    2DQuant {\textcolor{black!60}{(NeurIPS’24)}}   & 4 / 4 & 37.87 & 0.9594 & 33.41 & 0.9161 & 32.02 & 0.8971 & 31.84 & 0.9251 & 38.31 & 0.9761 \\
    FIMA-Q {\textcolor{black!60}{(CVPR’25)}}  & 4 / 4 & 37.71 & 0.9598 & 33.38 & 0.9160 & 32.01 & 0.8964 & 31.95 & 0.9283 & 38.44 & 0.9764 \\
    APHQ-ViT {\textcolor{black!60}{(CVPR’25)}} & 4 / 4 & 37.76 & 0.9591 & 33.41 & 0.9165 & 31.94 & 0.8965 & 31.98 & 0.9280 & 38.34 & 0.9759 \\
    \textbf{Ours} & \textbf{4 / 4}
      & \textbf{37.91} & \textbf{0.9600} & \textbf{33.47} & \textbf{0.9171} & \textbf{32.09} & \textbf{0.8978}
      & \textbf{32.06} & \textbf{0.9287} & \textbf{38.51} & \textbf{0.9773} \\
    \midrule
    2DQuant {\textcolor{black!60}{(NeurIPS’24)}}      & 2 / 2 & 36.00 & 0.9497 & 31.98 & 0.9012 & 30.91 & 0.8810 & 28.62 & 0.8819 & 34.40 & 0.9602 \\
    FIMA-Q {\textcolor{black!60}{(CVPR’25)}}  & 2 / 2 & 36.06 & 0.9515 & 32.10 & 0.9048 & 31.01 & 0.8848 & 28.77 & 0.8873 & 34.75 & 0.9638 \\
    APHQ-ViT {\textcolor{black!60}{(CVPR’25)}} & 2 / 2 & 36.14 & 0.9517 & 32.14 & 0.9049 & 31.04 & 0.8850 & 28.86 & 0.8885 & 34.99 & 0.9644 \\
    \textbf{Ours} & \textbf{2 / 2}
      & \textbf{36.58} & \textbf{0.9530} & \textbf{32.36} & \textbf{0.9056} & \textbf{31.17} & \textbf{0.8853}
      & \textbf{29.18} & \textbf{0.8908} & \textbf{35.47} & \textbf{0.9657} \\
    \bottomrule
  \end{tabular}
\end{table}

\begin{table}[H]
  \caption{$\times2$ SR on SwinIR-light under the Weight-training PTQ with LoRA protocol.}
  \label{tab:swinir_lora}
  \centering
  \scriptsize
  \setlength{\tabcolsep}{3pt}
  \renewcommand{\arraystretch}{0.90}
  \begin{tabular}{l c
  c@{\hspace{4pt}}c
  c@{\hspace{4pt}}c
  c@{\hspace{4pt}}c
  c@{\hspace{4pt}}c
  c@{\hspace{4pt}}c}
    \toprule
     \multirow{2}{*}{Method} & \multirow{2}{*}{W / A}
      & \multicolumn{2}{c}{Set5}
      & \multicolumn{2}{c}{Set14}
      & \multicolumn{2}{c}{BSD100}
      & \multicolumn{2}{c}{Urban100}
      & \multicolumn{2}{c}{Manga109} \\
    \cmidrule(lr){3-4} \cmidrule(lr){5-6} \cmidrule(lr){7-8} \cmidrule(lr){9-10} \cmidrule(lr){11-12}
     & & PSNR$\uparrow$ & SSIM$\uparrow$ & PSNR$\uparrow$ & SSIM$\uparrow$
         & PSNR$\uparrow$ & SSIM$\uparrow$ & PSNR$\uparrow$ & SSIM$\uparrow$
         & PSNR$\uparrow$ & SSIM$\uparrow$ \\
    \midrule
    SwinIR-light & 32 / 32 & 38.14 & 0.9611 & 33.86 & 0.9206 & 32.31 & 0.9012 & 32.76 & 0.9340 & 39.12 & 0.9783 \\
    \midrule
    SPR$^2$Q {\textcolor{black!60}{(ICLR’26)}} & 4 / 4 & 37.95 & 0.9602 & 33.55 & 0.9178 & 32.14 & 0.8989 & 32.04 & 0.9275 & 38.59 & 0.9769 \\
    \textbf{Ours + LoRA} & \textbf{4 / 4}
      & \textbf{38.04} & \textbf{0.9608} & \textbf{33.65} & \textbf{0.9193} & \textbf{32.22} & \textbf{0.9003}
      & \textbf{32.28} & \textbf{0.9302} & \textbf{38.75} & \textbf{0.9776} \\
    \midrule
    SPR$^2$Q {\textcolor{black!60}{(ICLR’26)}} & 2 / 2 & 37.28 & 0.9572 & 32.83 & 0.9113 & 31.63 & 0.8921 & 30.20 & 0.9077 & 37.08 & 0.9726 \\
    \textbf{Ours + LoRA} & \textbf{2 / 2}
      & \textbf{37.69} & \textbf{0.9596} & \textbf{33.22} & \textbf{0.9157} & \textbf{31.96} & \textbf{0.8971}
      & \textbf{31.16} & \textbf{0.9193} & \textbf{37.74} & \textbf{0.9753} \\
    \bottomrule
  \end{tabular}
\end{table}

\subsection{Vision Mamba on ImageNet}
\label{app:vim}

To test the observation outside restoration, we evaluate Vision
Mamba~\citep{visionmamba} on ImageNet classification. The observation of
Sec.~\ref{sec:obs} transfers. Token-range CoV increases after the Mamba block
in all 24 blocks of both models, with an average amplification of
$4.94\times$ on Vim-Tiny and $3.11\times$ on Vim-Small, which is larger than
the $1.80\times$ measured on MambaIRv2-Light.

For quantization we follow the scope and the 256-image calibration protocol of
PTQ4VM~\citep{ptq4vm} and apply our group-conditional quantizers with $K{=}8$.
The predictor runs once per block from the pre-scan block input, and the same
group indices are reused across all six per-token linear sites in the block.
Since classification outputs logits, we replace the pixel loss with a logit KD
loss and drop the FPR term while keeping feature distillation. Only the
activation clipping bounds are trained. Table~\ref{tab:vim} shows that ours is
best in every setting.

\begin{table}[H]
  \caption{ImageNet Top-1 accuracy in percent on Vision Mamba under PTQ.
  Calibration and quantization scope follow PTQ4VM.}
  \label{tab:vim}
  \centering
  \small
  \setlength{\tabcolsep}{8pt}
  \renewcommand{\arraystretch}{1.05}
  \begin{tabular}{l c cc}
    \toprule
    Method & W / A & Vim-Tiny & Vim-Small \\
    \midrule
    FP & 32 / 32 & 76.07 & 80.48 \\
    \midrule
    Layer-wise PTQ & 6 / 6 & 71.92 & 76.66 \\
    PTQ4VM {\textcolor{black!60}{(WACV’25)}} & 6 / 6 & 74.42 & 79.62 \\
    \textbf{Ours} & \textbf{6 / 6} & \textbf{75.88} & \textbf{79.95} \\
    \midrule
    Layer-wise PTQ & 4 / 4 & 53.61 & 67.06 \\
    PTQ4VM {\textcolor{black!60}{(WACV’25)}}& 4 / 4 & 56.62 & 70.41 \\
    \textbf{Ours} & \textbf{4 / 4} & \textbf{57.98} & \textbf{72.50} \\
    \bottomrule
  \end{tabular}
\end{table}

Although the token-range dispersion is even larger on Vim than in restoration,
its impact is task-dependent. Classification aggregates token features into a
single prediction, whereas restoration is dense per-pixel regression under
static deployment, where every token is on the critical path with all ranges
fixed ahead of time. The dispersion therefore becomes the first-order
bottleneck in restoration. At aggressive bit-widths it resurfaces in
classification as well, where ours again improves over PTQ4VM.

\subsection{Larger MambaIR Backbones}
\label{app:base}

Table~\ref{tab:mambair_base} evaluates MambaIRv1-Base and MambaIRv2-Base at
$\times2$ under the Weight-freeze protocol with the same quantization scope,
calibration setup, and baselines as Appendix~\ref{app:additional_results}.
Ours is best in every setting, so the failure mode identified in
Sec.~\ref{sec:obs} and its remedy are not confined to lightweight models.

\begin{table}[H]
  \caption{$\times2$ SR on MambaIRv1-Base and MambaIRv2-Base under the
  Weight-freeze PTQ protocol.}
  \label{tab:mambair_base}
  \centering
  \footnotesize
  \setlength{\tabcolsep}{6pt}
  \renewcommand{\arraystretch}{1.05}
  \begin{tabular}{l l c cc cc}
    \toprule
    \multirow{2}{*}{Backbone} & \multirow{2}{*}{Method} & \multirow{2}{*}{W / A}
      & \multicolumn{2}{c}{Set14} & \multicolumn{2}{c}{Urban100} \\
    \cmidrule(lr){4-5} \cmidrule(lr){6-7}
     & & & PSNR$\uparrow$ & SSIM$\uparrow$ & PSNR$\uparrow$ & SSIM$\uparrow$ \\
    \midrule
    \multirow{7}{*}{MambaIRv2-Base}
      & FP             & 32 / 32 & 34.89 & 0.9275 & 34.49 & 0.9468 \\
    \cmidrule(lr){2-7}
      & Layer-wise PTQ & 4 / 4 & 32.39 & 0.9060 & 29.25 & 0.8930 \\
      & OAQ-SR {\textcolor{black!60}{(ICCV'25)}}         & 4 / 4 & 33.04 & 0.9132 & 31.25 & 0.9195 \\
      & \textbf{Ours}  & \textbf{4 / 4} & \textbf{33.28} & \textbf{0.9157} & \textbf{31.60} & \textbf{0.9236} \\
    \cmidrule(lr){2-7}
      & Layer-wise PTQ & 2 / 2 & 30.54 & 0.8556 & 27.53 & 0.8398 \\
      & OAQ-SR {\textcolor{black!60}{(ICCV'25)}}         & 2 / 2 & 31.33 & 0.8958 & 27.81 & 0.8699 \\
      & \textbf{Ours}  & \textbf{2 / 2} & \textbf{31.47} & \textbf{0.8974} & \textbf{28.13} & \textbf{0.8756} \\
    \midrule
    \multirow{7}{*}{MambaIRv1-Base}
      & FP             & 32 / 32 & 34.67 & 0.9261 & 34.15 & 0.9446 \\
    \cmidrule(lr){2-7}
      & Layer-wise PTQ & 4 / 4 & 31.72 & 0.8808 & 29.07 & 0.8719 \\
      & OAQ-SR {\textcolor{black!60}{(ICCV'25)}}         & 4 / 4 & 32.19 & 0.8952 & 29.53 & 0.8874 \\
      & \textbf{Ours}  & \textbf{4 / 4} & \textbf{33.07} & \textbf{0.9094} & \textbf{31.17} & \textbf{0.9153} \\
    \cmidrule(lr){2-7}
      & Layer-wise PTQ & 2 / 2 & 29.13 & 0.7846 & 25.47 & 0.6770 \\
      & OAQ-SR {\textcolor{black!60}{(ICCV'25)}}         & 2 / 2 & 30.84 & 0.8830 & 27.43 & 0.8561 \\
      & \textbf{Ours}  & \textbf{2 / 2} & \textbf{31.23} & \textbf{0.8940} & \textbf{27.71} & \textbf{0.8672} \\
    \bottomrule
  \end{tabular}
\end{table}

\subsection{Relation to Channel-Axis Quantization Transforms}
\label{app:transforms}

Rotation, affine, and scaling quantizers from the LLM and diffusion
literature~\citep{ashkboos2024quarot,liu2025spinquant,ma2024affinequant,sun2024flatquant,xiao2023smoothquant,li2024svdquant}
share one property with prior SR and Mamba PTQ methods. The quantization range
is assigned independently of token identity, through a channel-axis transform
shared by every token. This is sound when token distributions are homogeneous,
which is exactly the assumption that selective scan breaks in
Sec.~\ref{sec:obs}. To verify this empirically, we ported the core transforms
onto MambaIRv2-Light under the identical protocol, sites, calibration set, and
budget. Hadamard covers QuaRot~\citep{ashkboos2024quarot},
SpinQuant~\citep{liu2025spinquant} covers learned rotation, and Affine covers
AffineQuant~\citep{ma2024affinequant} and FlatQuant~\citep{sun2024flatquant},
with SmoothQuant-style scale migration~\citep{xiao2023smoothquant} as its
diagonal case. Atom~\citep{zhao2024atom} alone acts on the token axis but
through runtime statistics, which C1 rules out, so ours is its static
counterpart.

Table~\ref{tab:transforms} shows two findings. First, even the best channel-axis
transform on each benchmark trails ours by up to 2.12~dB at W4A4. Second, the
two axes compose. SpinQuant combined with our method is best in every setting,
with the largest gains at W2A2. The two remove different error sources, so
Token-Group Quantization occupies a design axis these literatures do not cover
while remaining compatible with them.

\begin{table}[H]
  \caption{$\times2$ SR on MambaIRv2-Light with channel-axis transforms ported
  onto the identical Weight-freeze PTQ protocol. Affine covers AffineQuant and
  FlatQuant, and Hadamard covers QuaRot.}
  \label{tab:transforms}
  \centering
  \scriptsize
  \setlength{\tabcolsep}{3pt}
  \renewcommand{\arraystretch}{0.90}
  \begin{tabular}{l c
  c@{\hspace{4pt}}c
  c@{\hspace{4pt}}c
  c@{\hspace{4pt}}c
  c@{\hspace{4pt}}c
  c@{\hspace{4pt}}c}
    \toprule
     \multirow{2}{*}{Method} & \multirow{2}{*}{W / A}
      & \multicolumn{2}{c}{Set5}
      & \multicolumn{2}{c}{Set14}
      & \multicolumn{2}{c}{BSD100}
      & \multicolumn{2}{c}{Urban100}
      & \multicolumn{2}{c}{Manga109} \\
    \cmidrule(lr){3-4} \cmidrule(lr){5-6} \cmidrule(lr){7-8} \cmidrule(lr){9-10} \cmidrule(lr){11-12}
     & & PSNR$\uparrow$ & SSIM$\uparrow$ & PSNR$\uparrow$ & SSIM$\uparrow$
         & PSNR$\uparrow$ & SSIM$\uparrow$ & PSNR$\uparrow$ & SSIM$\uparrow$
         & PSNR$\uparrow$ & SSIM$\uparrow$ \\
    \midrule
    MambaIRv2-Light & 32 / 32 & 38.26 & 0.9615 & 34.09 & 0.9221 & 32.36 & 0.9019 & 33.26 & 0.9378 & 39.35 & 0.9785 \\
    \midrule
    Affine    & 4 / 4 & 35.80 & 0.9452 & 32.81 & 0.9091 & 30.89 & 0.8789 & 30.57 & 0.9163 & 34.67 & 0.9550 \\
    Hadamard  & 4 / 4 & 36.33 & 0.9492 & 32.61 & 0.9085 & 31.04 & 0.8823 & 30.36 & 0.9123 & 35.73 & 0.9655 \\
    SpinQuant & 4 / 4 & 36.42 & 0.9500 & 32.50 & 0.9063 & 31.08 & 0.8828 & 30.74 & 0.9188 & 35.54 & 0.9622 \\
    Ours      & 4 / 4 & 37.56 & 0.9578 & 33.24 & 0.9149 & 31.86 & 0.8950 & 31.34 & 0.9204 & 37.85 & 0.9742 \\
    \textbf{SpinQuant + Ours} & \textbf{4 / 4}
      & \textbf{37.69} & \textbf{0.9588} & \textbf{33.40} & \textbf{0.9163} & \textbf{31.96} & \textbf{0.8965}
      & \textbf{31.61} & \textbf{0.9228} & \textbf{37.94} & \textbf{0.9750} \\
    \midrule
    Affine    & 2 / 2 & 34.59 & 0.9365 & 31.18 & 0.8904 & 30.32 & 0.8703 & 27.71 & 0.8643 & 32.31 & 0.9438 \\
    Hadamard  & 2 / 2 & 34.72 & 0.9389 & 31.23 & 0.8923 & 30.37 & 0.8722 & 27.75 & 0.8656 & 32.35 & 0.9456 \\
    SpinQuant & 2 / 2 & 34.84 & 0.9400 & 31.29 & 0.8935 & 30.42 & 0.8734 & 27.79 & 0.8668 & 32.46 & 0.9467 \\
    Ours      & 2 / 2 & 35.02 & 0.9436 & 31.41 & 0.8966 & 30.51 & 0.8766 & 27.90 & 0.8708 & 32.67 & 0.9508 \\
    \textbf{SpinQuant + Ours} & \textbf{2 / 2}
      & \textbf{36.23} & \textbf{0.9513} & \textbf{32.17} & \textbf{0.9037} & \textbf{31.06} & \textbf{0.8841}
      & \textbf{28.77} & \textbf{0.8850} & \textbf{34.72} & \textbf{0.9614} \\
    \bottomrule
  \end{tabular}
\end{table}

\section{Rationale for Quantization-Aware Descriptors}
\label{app:descriptor}

\subsection{Why These Descriptor Statistics?}
\label{app:descriptor_rationale}

We motivate the descriptor in Eq.~\ref{eq:phi} from the error decomposition of
a uniform quantizer. For clarity, consider symmetric clipping $[-c,c]$ for a
token-level random variable $X_i$ drawn from the channel entries of
$x_i^{(\ell,m)}$. The quantization step size is
$\Delta(c)=2c/(2^b-1)$. Under the standard high-resolution
approximation, the per-token quantization error can be written as
\begin{equation}
E_i(c) \approx
\frac{\Delta(c)^2}{12}\Pr(|X_i|\le c)
+
\mathbb{E}\left[(|X_i|-c)^2_+\right].
\label{eq:descriptor_rationale}
\end{equation}
Here, $(z)_+=\max(z,0)$. The first term is the rounding error within the
clipping range and increases with the step size, while the second term is the
clipping error from tail activations. Therefore, tokens should share a clipping
bound only when their bulk scale and tail behavior are similar.

The descriptor in Eq.~\ref{eq:phi} provides a compact proxy for this
precision--clipping trade-off. $\mathrm{RMS}(x_i^{(\ell,m)})$ captures the
bulk activation energy that influences rounding sensitivity, $q_{0.99}$ and
$q_{0.999}$ capture near-tail and extreme-tail scales, and $\max |x_i^{(\ell,m)}|$
captures the worst-case tail extent. Although these statistics are not sufficient
statistics for the full activation distribution, they characterize the dominant
factors that shape the quantization-error curve over clipping bounds. Clustering
tokens in this descriptor space therefore groups tokens that are likely to have
similar MSE-preferred clipping ranges.

\section{Pre-scan Predictor Details and Robustness}
\label{app:predictor}

\subsection{Prediction Accuracy and Error Structure}

The pre-scan predictor is trained to reproduce pseudo-labels obtained by
$K$-means clustering in the downstream activation-descriptor space. For $K=8$,
the predictor achieves approximately \textbf{84\%} top-1 accuracy on held-out
tokens. This accuracy should not be interpreted as classification accuracy
against semantic ground truth. The pseudo-labels are calibration artifacts that
compress quantization-error regimes, so the accuracy measures how well
pre-scan features recover the descriptor-space partition used for grouping.

More importantly, prediction errors are structured. We further observe that
many misclassified tokens are assigned to nearby clusters in the descriptor
space rather than to distant regimes. Since the descriptor is constructed from
statistics that dominate the rounding--clipping trade-off, nearby clusters tend
to have similar MSE-preferred clipping bounds. Therefore, such errors still
often select bounds that are close to the oracle-preferred range. This
supports the robustness of using a lightweight pre-scan predictor instead of
runtime post-scan group assignment.

The pseudo-labels should therefore be understood not as semantic labels, but as
a compact representation of quantization-error regimes. What matters is not
perfect label stability, but whether tokens assigned to the same group exhibit
similar clipping-error and rounding-error behavior.

\subsection{Robustness to Inference-Time Predictor Errors}
\label{app:predictor_robustness}

While Table~\ref{tab:grouping_rule_main} compares different grouping
rules---each requiring its own calibration---we now ask a complementary
question: how much does the framework tolerate \emph{runtime} errors in the
predictor output? In Table~\ref{tab:random_grouping_ablation}, we keep the
calibrated bound tables and the trained predictor exactly as in our default
deployment, and only at evaluation time randomly reassign a fraction $\rho$
of the predicted group ids to a uniformly random group in
$\{1,\dots,K\}$. No re-calibration, no bound re-refinement, and no predictor
re-training is performed. The $\rho{=}0\%$ row therefore matches our default
result, while $\rho{=}100\%$ corresponds to fully random group ids \emph{on
top of} the bound tables calibrated for our predictor, and is conceptually
distinct from the ``Random groups'' row of
Table~\ref{tab:grouping_rule_main}, where bounds are calibrated jointly
with random grouping.

\begin{table}[H]
  \centering
  \caption{
  \textbf{Robustness to inference-time predictor perturbation} on
  MambaIRv2-Light ($\times2$, W4A4) under the Weight-freeze PTQ protocol.
  We use the calibrated bound tables and the trained predictor from our
  default setting and, only at evaluation time, randomly reassign a fraction
  $\rho$ of predicted group ids to a uniformly random group in
  $\{1,\dots,K\}$. Calibration, bound refinement, and the predictor are
  unchanged across rows. The $\rho{=}0\%$ row matches our default deployment.
  }
  \label{tab:random_grouping_ablation}
  \small
  \setlength{\tabcolsep}{7pt}
  \renewcommand{\arraystretch}{1.12}
  \begin{tabular}{l cc cc}
    \toprule
    \multirow{2}{*}{Perturbation rate $\rho$}
    & \multicolumn{2}{c}{Set14}
    & \multicolumn{2}{c}{Urban100} \\
    \cmidrule(lr){2-3} \cmidrule(lr){4-5}
    & PSNR$\uparrow$ & SSIM$\uparrow$
    & PSNR$\uparrow$ & SSIM$\uparrow$ \\
    \midrule
    $\rho = 0\%$ (ours, no perturbation)
      & 33.24 & 0.9149
      & 31.34 & 0.9204 \\
    $\rho = 5\%$
      & 33.21 & 0.9145
      & 31.34 & 0.9204 \\
    $\rho = 10\%$
      & 33.18 & 0.9143
      & 31.33 & 0.9202 \\
    $\rho = 20\%$
      & 33.09 & 0.9137
      & 31.25 & 0.9193 \\
    $\rho = 30\%$
      & 33.01 & 0.9127
      & 31.13 & 0.9178 \\
    $\rho = 50\%$
      & 32.72 & 0.9108
      & 31.02 & 0.9165 \\
    $\rho = 100\%$ (uniform random ids)
      & 32.56 & 0.9084
      & 30.45 & 0.9121 \\
    \bottomrule
  \end{tabular}
\end{table}

The empirical predictor error rate is approximately $16\%$ ($K{=}8$, top-1
accuracy $84\%$). Around this regime, perturbation has only a marginal
effect: at $\rho{=}10\%$ and $\rho{=}20\%$, the Set14 PSNR drops by only
$0.06$ dB and $0.15$ dB from the unperturbed default, with similarly small
drops on Urban100. Even at $\rho{=}50\%$---where half of all predicted
group ids are randomly corrupted---PSNR degrades by only $0.52$ dB on
Set14 and $0.32$ dB on Urban100, well below the gap between layer-wise
PTQ and our method (Table~\ref{tab:grouping_rule_main}).

A particularly informative comparison is the $\rho{=}100\%$ row, in which
predicted group ids are entirely replaced by uniformly random ids while
the bound tables remain those calibrated for our predictor. This setting
still attains $32.56$ / $30.45$ dB on Set14 / Urban100, clearly above the
``Random groups'' row of Table~\ref{tab:grouping_rule_main}
($31.76$ / $29.91$ dB), where the bound tables themselves were calibrated
under random grouping. The $0.80$ / $0.54$ dB gap indicates that a
substantial portion of the benefit comes from the calibrated bound tables
themselves, with the predictor providing additional gains on top.

Together with the structured-error analysis in Appendix~\ref{app:predictor},
this confirms that the framework does not rely on the predictor being
near-perfect. Small fluctuations in predicted group ids are absorbed by
the bound tables, and the method degrades gracefully rather than
catastrophically as prediction quality decreases.

\section{Additional Implementation Details}
\label{app:implementation_details}

\subsection{Pre-scan Predictor Training Details}
\label{app:predictor_details}

The pre-scan predictor is a shared two-layer FFN
($C{\rightarrow}64{\rightarrow}K$ with GELU, where $C{=}48$ for
MambaIRv2-Light and $K{=}8$ by default). It is trained only during calibration
to reproduce the descriptor-K-means pseudo-labels using cross-entropy. We use
Adam with learning rate $10^{-3}$, batch size 1024, 20 epochs, an 80/20
train/validation split, and random seed 0. The input is the pre-ASSM feature at
each token--layer pair, and the target is the $K$-means label obtained from
descriptors concatenated over all eight group-conditional sites: three
downstream linear sites and five scan-internal per-token operands. After
training, the FFN is frozen and used only to assign group indices for fixed
clipping-bound table lookup.

\subsection{Quantization Scope}
\label{app:quant_scope}

We apply low-bit quantization to two functional groups of activations. First,
we quantize the inputs to the linear and pointwise convolution modules in the
backbone, including the post-scan projection \texttt{out\_proj} and the two
\texttt{convffn2} linear layers in each ASSM block. This part accounts for
$85.9\%$ of the FLOPs. Second, inside selective scan, we quantize five
token-adaptive operands: \texttt{xs}, \texttt{dts}, \texttt{delta},
\texttt{Bs}, and \texttt{Cs}. These scan operands account for an additional
$4.9\%$ of the FLOPs. Together, the low-bit path covers $90.8\%$ of the total
FLOPs, matching the quantization scope reported in Table~\ref{tab:flops_scope}.

Among these quantized activations, group-wise clipping is applied to the eight
token-indexed sites used for pseudo-label construction: \texttt{out\_proj},
\texttt{convffn2.fc1}, \texttt{convffn2.fc2}, and the five scan operands
\texttt{xs}, \texttt{dts}, \texttt{delta}, \texttt{Bs}, and \texttt{Cs}. The
remaining modules are kept in full precision due to sensitivity or limited
benefit from low-bit quantization.

For feature distillation, $\mathcal{H}$ contains the outputs of all
ASSM blocks. In MambaIRv2-Light, $|\mathcal{H}|=20$, and
$F_h,\hat{F}_h$ denote the corresponding full-precision teacher and quantized
student features after the residual output of block $h$.

\subsection{Reproducibility}
\label{app:reproducibility}

The source code is publicly available at \url{https://github.com/imaboybut/TGQ-MambaSR}.

\section{Robustness Analyses}
\label{app:robustness}

\subsection{Sensitivity to Calibration Sampling and Random Seeds}
\label{app:seeds}

The calibration pipeline involves patch sampling, $K$-means initialization,
predictor training, and bound refinement. We repeat the full pipeline from
scratch under five random seeds on MambaIRv2-Light at $\times2$. Each seed
jointly re-samples the 96 patches from DF2K, re-initializes the $K$-means, and
re-seeds predictor training. Bound initialization and refinement are re-run
per seed, so nothing is shared except the pretrained backbone.

The results are highly stable. The standard deviation across seeds stays below
0.03~dB PSNR and 0.005 SSIM on all five benchmarks at W4A4, and below 0.05~dB
and 0.01 at W2A2. These deviations are far smaller than the margins over the
strongest baseline, such as the 0.85~dB gain over OAQ-SR on Urban100 at W4A4,
so the improvements are not a favorable calibration draw. The predictor
reaches $85\pm1.3\%$ top-1 accuracy across runs, close to the 84\% reported in
Appendix~\ref{app:predictor}, which indicates that the pseudo-labels remain
similarly learnable across seeds. This stability is consistent with the
design, since the bound tables absorb group-assignment fluctuations and such
small swings are well within the tolerance shown in
Table~\ref{tab:random_grouping_ablation}.


\subsection{Robustness to Input Resolution}
\label{app:resolution}

Calibration uses $64\times64$ LR crops of about 4k tokens, while selective
scan runs unwindowed over $L{=}HW$, so Urban100 evaluation already reaches
about 308k tokens, which is 75 times the calibration length. To isolate
resolution from content, Table~\ref{tab:resolution} freezes the predictor and
bound tables and crops the same Urban100 images across a 64 times token range.
We report the gap between FP and Ours, since crop size shifts content
statistics for both models identically and the gap isolates what the frozen
grouping rule loses. The gap remains nearly flat across the whole range, and
at 16 times the calibration length it is no larger than at the calibration
length itself. The method is therefore insensitive to input resolution, and
this flatness under fixed $K{=}8$ suggests that per-resolution retuning of $K$
is unnecessary.

\begin{table}[H]
  \caption{Resolution robustness on Urban100 at $\times2$ and W4A4 over 20
  images. Predictor and bound tables are frozen and only the LR input crop
  size varies. Gap is FP minus Ours.}
  \label{tab:resolution}
  \centering
  \small
  \setlength{\tabcolsep}{6pt}
  \renewcommand{\arraystretch}{1.05}
  \begin{tabular}{c c c cc cc cc}
    \toprule
    \multirow{2}{*}{LR crop} & \multirow{2}{*}{Tokens} & \multirow{2}{*}{Ratio to calib.}
      & \multicolumn{2}{c}{FP} & \multicolumn{2}{c}{Ours} & \multicolumn{2}{c}{Gap} \\
    \cmidrule(lr){4-5} \cmidrule(lr){6-7} \cmidrule(lr){8-9}
      & & & PSNR & SSIM & PSNR & SSIM & PSNR & SSIM \\
    \midrule
    $32\times32$   & 1{,}024  & 0.25$\times$ & 33.77 & 0.9127 & 31.88 & 0.8929 & 1.89 & 0.0198 \\
    $64\times64$   & 4{,}096  & 1$\times$    & 32.60 & 0.9084 & 30.48 & 0.8825 & 2.12 & 0.0259 \\
    $128\times128$ & 16{,}384 & 4$\times$    & 31.89 & 0.9105 & 29.78 & 0.8873 & 2.11 & 0.0232 \\
    $256\times256$ & 65{,}536 & 16$\times$   & 32.13 & 0.9217 & 30.07 & 0.9003 & 2.06 & 0.0214 \\
    \bottomrule
  \end{tabular}
\end{table}

\subsection{Number of Token Groups on a Larger Backbone}
\label{app:k_base}

Table~\ref{tab:k_base} repeats the $K$ sweep of Table~\ref{tab:k_ablation} on
MambaIRv2-Base, extended to $K{=}12$ and $K{=}16$. The optimum shifts mildly
upward with model size, since a wider channel dimension supports more
distinguishable activation regimes, but the returns diminish quickly. Moving
from $K{=}1$ to $K{=}8$ already recovers 89\% of the total range on Set14.
$K{=}8$ therefore remains a robust default without per-model tuning, and
larger backbones can raise $K$ for a small further gain. The identical
two-layer FFN transfers without architectural change, and only its input width
follows the backbone channel dimension.

\begin{table}[H]
  \caption{Effect of the number of token groups $K$ on MambaIRv2-Base at
  $\times2$ and W4A4.}
  \label{tab:k_base}
  \centering
  \small
  \setlength{\tabcolsep}{8pt}
  \renewcommand{\arraystretch}{1.05}
  \begin{tabular}{c cc cc}
    \toprule
    \multirow{2}{*}{$K$} & \multicolumn{2}{c}{Set14} & \multicolumn{2}{c}{Urban100} \\
    \cmidrule(lr){2-3} \cmidrule(lr){4-5}
      & PSNR$\uparrow$ & SSIM$\uparrow$ & PSNR$\uparrow$ & SSIM$\uparrow$ \\
    \midrule
    1  & 32.55 & 0.9099 & 30.98 & 0.9084 \\
    4  & 33.20 & 0.9141 & 31.37 & 0.9193 \\
    8, default & 33.28 & 0.9157 & 31.60 & 0.9236 \\
    12 & 33.34 & 0.9163 & 31.68 & 0.9244 \\
    16 & \textbf{33.37} & \textbf{0.9168} & \textbf{31.71} & \textbf{0.9249} \\
    \bottomrule
  \end{tabular}
\end{table}

\section{Deployment Details}
\label{app:deployment}

We implement a CUDA deployment path for the W4A4 quantized model to evaluate
practical 4-bit acceleration. After calibration, all clipping-bound tables,
quantization scales, weight quantization parameters, and pre-scan predictor
parameters are frozen. Runtime execution predicts token groups from pre-scan
features and selects the corresponding fixed clipping bounds, without runtime
statistics collection, dynamic clipping updates, or post-scan group assignment.

For grouped activation sites, calibration-time fake-quantization modules are
replaced with static CUDA kernels that fuse group-wise bound lookup, clamping,
rounding, and dequantization. The activation and weight quantization codes are
constrained to the 4-bit range, while byte-aligned integer buffers are used for
efficient CUDA execution. Quantized linear layers use pre-quantized weights,
integer CUDA kernels, and dequantization with fixed activation and weight
quantization parameters.

We also apply CUDA graph capture after static conversion to reduce CPU dispatch
overhead. As reported in Table~\ref{tab:latency_cost}, the W4A4 deployment
achieves 1.70$\times$ speedup on A6000 and 2.23$\times$ speedup on Jetson Orin
Nano over FP16, demonstrating practical 4-bit acceleration for Token-Group
Quantization.

We measure end-to-end latency on Set5 with batch size 1 using the same
evaluation script, backend, graph mode, and timing protocol for FP16, the
4-bit base path, and our grouped 4-bit path. For each image, we run 5 warm-up
iterations, synchronize the device, run 10 timed repetitions, synchronize again,
and report the average latency per image. Since Set5 contains images with
different resolutions, CUDA graph capture is performed once for each encountered
input shape and then reused for that shape. The reported latency excludes graph
construction time but includes pre-scan predictor execution, group-wise bound
lookup, quantization/dequantization, and low-bit linear execution.

\paragraph{Overhead breakdown.}
The measured latency in Table~\ref{tab:latency_cost} includes every component
of the grouped path. Predictor execution is inside the timed path, and
group-wise bound lookup, clamping, rounding, and dequantization are fused into
the static CUDA kernels described above. The resulting overhead over the base
4-bit path is 3.4\% on A6000 and 2.2\% on Jetson Orin Nano, and the added
memory is about 5\,KB of bound tables plus a two-layer FFN, so the two paths
are nearly identical in cost. By hardware-friendly we mean the property that
matters for static deployment. There are no runtime statistics and no dynamic
clipping updates, and all parameters are fixed after calibration, so the path
stays compatible with ahead-of-time compilation and CUDA graph capture.

\paragraph{Runtime post-scan oracle.}
The 4-bit Oracle rows of Table~\ref{tab:latency_cost} measure the most
favorable dynamic implementation of the post-scan oracle of
Table~\ref{tab:grouping_rule_main}. Descriptor statistics of Eq.~\ref{eq:phi}
are computed by dedicated kernels and groups are assigned by nearest-centroid
lookup against the offline $K$-means centroids, with no $K$-means re-run and
with the same CUDA graph capture and timing protocol. Its cost has three
sources. First, and dominant, the block must run twice. The group id of a
token is determined by statistics of its post-scan activations at the eight
grouped sites, but those activations exist only after the block has executed,
and the quantizers of the same block already need the id. The only resolution
is to run the block once to observe, assign groups, and run it again
quantized. Our predictor breaks this dependency by reading pre-scan features,
so one pass suffices. Second, the descriptors themselves are costly. At 160
sites, eight per block over 20 blocks, the activation tensor is read once more
to obtain per-token $q_{0.99}$, $q_{0.999}$, RMS, and max. The two quantiles
are order statistics that cannot be fused into the quantization kernel, so
this step adds separate memory-bound kernels and a synchronization point
between the two passes. Third, nearest-centroid assignment over the
32-dimensional descriptor is cheap but adds one more launch per block.
As a result, the oracle takes 254.5\,ms on A6000 and 3213.9\,ms on Jetson Orin
Nano, a 0.71$\times$ and 0.98$\times$ speedup over FP16, so 4-bit execution
brings no acceleration at all, whereas the base path and ours reach
1.76$\times$ and 1.70$\times$ on A6000 and 2.28$\times$ and 2.23$\times$ on
Orin Nano. The oracle's 0.17\,dB advantage in Table~\ref{tab:grouping_rule_main}
therefore comes at a cost that static low-bit deployment cannot pay, whereas
ours retains most of that gain with a single pass and fixed bound tables.

\section{Frequency-Domain Evidence for FPR}
\label{app:freq_evidence}

Table~\ref{tab:component_ablation} in the main paper shows that the amplitude
and phase terms each help alone and are complementary in combination. Here we
show where in the spectrum the gain arises. Table~\ref{tab:bandwise}
decomposes the error of the quantized output against the FP output on
Urban100 at $\times2$ and W4A4 into low, mid, and high radial frequency bands
and reports the mean amplitude and phase error per band. Set14 shows the same
pattern. The variants correspond to the last four rows of
Table~\ref{tab:component_ablation}, and the row labeled none is
$\mathcal{L}_\mathrm{pix}{+}\mathcal{L}_\mathrm{feat}$ with token groups.

\begin{table}[H]
  \caption{Band-wise Fourier errors of the quantized output against the FP
  output on Urban100 at $\times2$ and W4A4. Lower is better. All variants use
  token groups, and none denotes
  $\mathcal{L}_\mathrm{pix}{+}\mathcal{L}_\mathrm{feat}$ without any
  frequency term.}
  \label{tab:bandwise}
  \centering
  \small
  \setlength{\tabcolsep}{7pt}
  \renewcommand{\arraystretch}{1.05}
  \begin{tabular}{l ccc ccc}
    \toprule
    \multirow{2}{*}{Frequency term}
      & \multicolumn{3}{c}{Amplitude error} & \multicolumn{3}{c}{Phase error} \\
    \cmidrule(lr){2-4} \cmidrule(lr){5-7}
      & Low & Mid & High & Low & Mid & High \\
    \midrule
    none             & 0.0208 & 0.2853 & 0.536 & 0.0213 & 0.3211 & 0.673 \\
    + amplitude only & 0.0206 & 0.2784 & \textbf{0.512} & 0.0212 & 0.3191 & 0.668 \\
    + phase only     & 0.0213 & 0.2830 & 0.531 & 0.0216 & 0.3181 & \textbf{0.635} \\
    + $\mathcal{L}_\mathrm{FPR}$, Ours & 0.0214 & 0.2850 & 0.518 & 0.0219 & 0.3213 & 0.643 \\
    \bottomrule
  \end{tabular}
\end{table}

Three observations follow. First, the error concentrates in the high band,
while low-band errors are an order of magnitude smaller and flat across
variants, so pixel and feature matching already cover low frequencies and
under-supervise high ones. Second, in the high band the amplitude term
primarily reduces amplitude error and the phase term primarily reduces phase
error, so the two emphasize complementary error axes. Third, each single term
concentrates its supervision on one axis, whereas FPR reduces both high-band
errors jointly. Since restoration quality depends on the joint error, only
FPR reaches the best PSNR and SSIM in Table~\ref{tab:component_ablation}.

Existing frequency-domain losses~\citep{fffloss,fffloss2} train network
weights and are supervised by ground-truth HR images or a larger teacher. FPR
differs on both sides. Only the clipping bounds are trained while all weights
stay frozen, and the teacher is the FP model supervising its own quantized
counterpart. The objective is thus not to learn a representation but to
correct the spectral error that quantization itself introduces.

\section{FPR Objective Weight Sensitivity}
\label{app:fpr_ablation}

Table~\ref{tab:fpr_lambda_ablation} studies the sensitivity to
$\lambda_\mathrm{FPR}$ in Eq.~\ref{eq:total_loss} under the
Weight-freeze PTQ protocol at $\times2$, W4A4. All other settings, including
the pre-scan predictor, bound initialization, $\lambda_\mathrm{feat}{=}5$,
and the bound-refinement schedule, are kept at their default values.
A small weight ($0.005$) provides too weak a frequency-domain signal,
while a large weight ($0.1$) competes with
$\mathcal{L}_\mathrm{pix}$ and $\mathcal{L}_\mathrm{feat}$ and degrades
high-frequency textures, as reflected by the larger drop on Urban100. The
performance is stable in a broad neighborhood around the default
$\lambda_\mathrm{FPR}{=}0.01$.

\begin{table}[H]
  \centering
  \caption{
  \textbf{Sensitivity to FPR objective weight $\lambda_\mathrm{FPR}$}
  on MambaIRv2-Light ($\times2$, W4A4) under the Weight-freeze PTQ protocol.
  All other settings are kept at their default values.
  $\lambda_\mathrm{FPR}{=}0.01$ corresponds to our default setting.
  }
  \label{tab:fpr_lambda_ablation}
  \small
  \setlength{\tabcolsep}{7pt}
  \renewcommand{\arraystretch}{1.12}
  \begin{tabular}{l cc cc}
    \toprule
    \multirow{2}{*}{$\lambda_\mathrm{FPR}$}
    & \multicolumn{2}{c}{Set14}
    & \multicolumn{2}{c}{Urban100} \\
    \cmidrule(lr){2-3} \cmidrule(lr){4-5}
    & PSNR$\uparrow$ & SSIM$\uparrow$
    & PSNR$\uparrow$ & SSIM$\uparrow$ \\
    \midrule
    $0.005$
      & 33.11 & 0.9124
      & 31.19 & 0.9184 \\
    $0.01$ (ours)
      & 33.24 & 0.9149
      & 31.34 & 0.9204 \\
    $0.05$
      & 33.22 & 0.9140
      & 31.31 & 0.9197 \\
    $0.1$
      & 33.09 & 0.9116
      & 31.04 & 0.9175 \\
    \bottomrule
  \end{tabular}
\end{table}

\clearpage
\section{Clipping-Scale Regimes Across Layers}
\label{app:clip_regime_layers}

To complement Fig.~\ref{fig:group_regime}, we visualize token-level preferred
clipping scales $c_i^\star$ across additional ASSM layers and post-scan sites.
These plots show that downstream descriptor groups often reveal stronger
clipping-scale structure than random groups, while predicted groups provide a
deployable but less sharply separated approximation.

\begin{figure}[H]
  \centering
  \includegraphics[width=\textwidth]{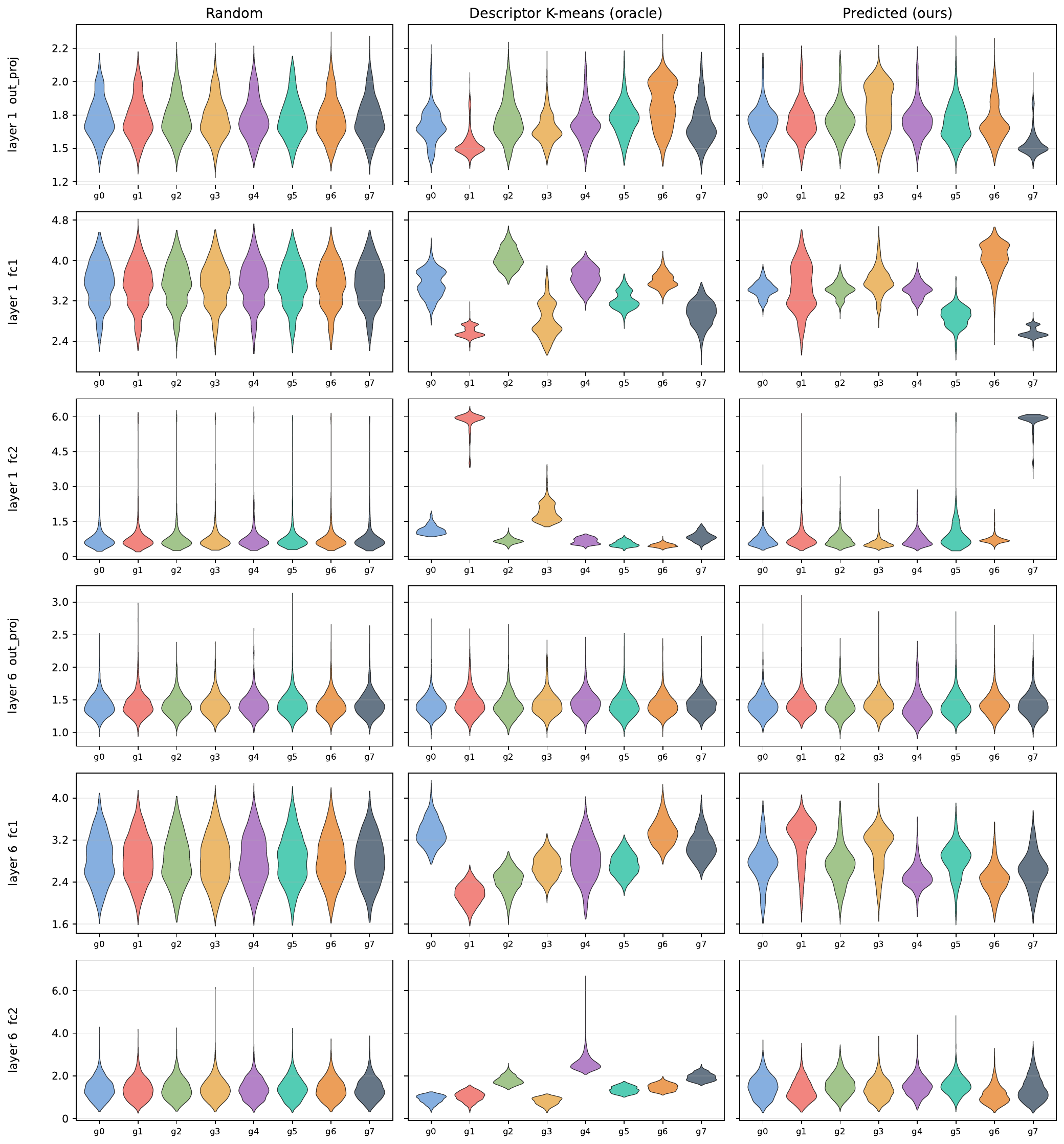}
  \caption{
  \textbf{Clipping-scale regimes in early/middle layers.}
  Token-level preferred clipping scales $c_i^\star$ for layers 1 and 6 under
  random, post-scan descriptor K-means (oracle), and predicted groups.
  }
  \label{fig:app_clip_regime_l1_l6}
\end{figure}

\begin{figure*}[!t]
  \centering
  \includegraphics[width=\textwidth]{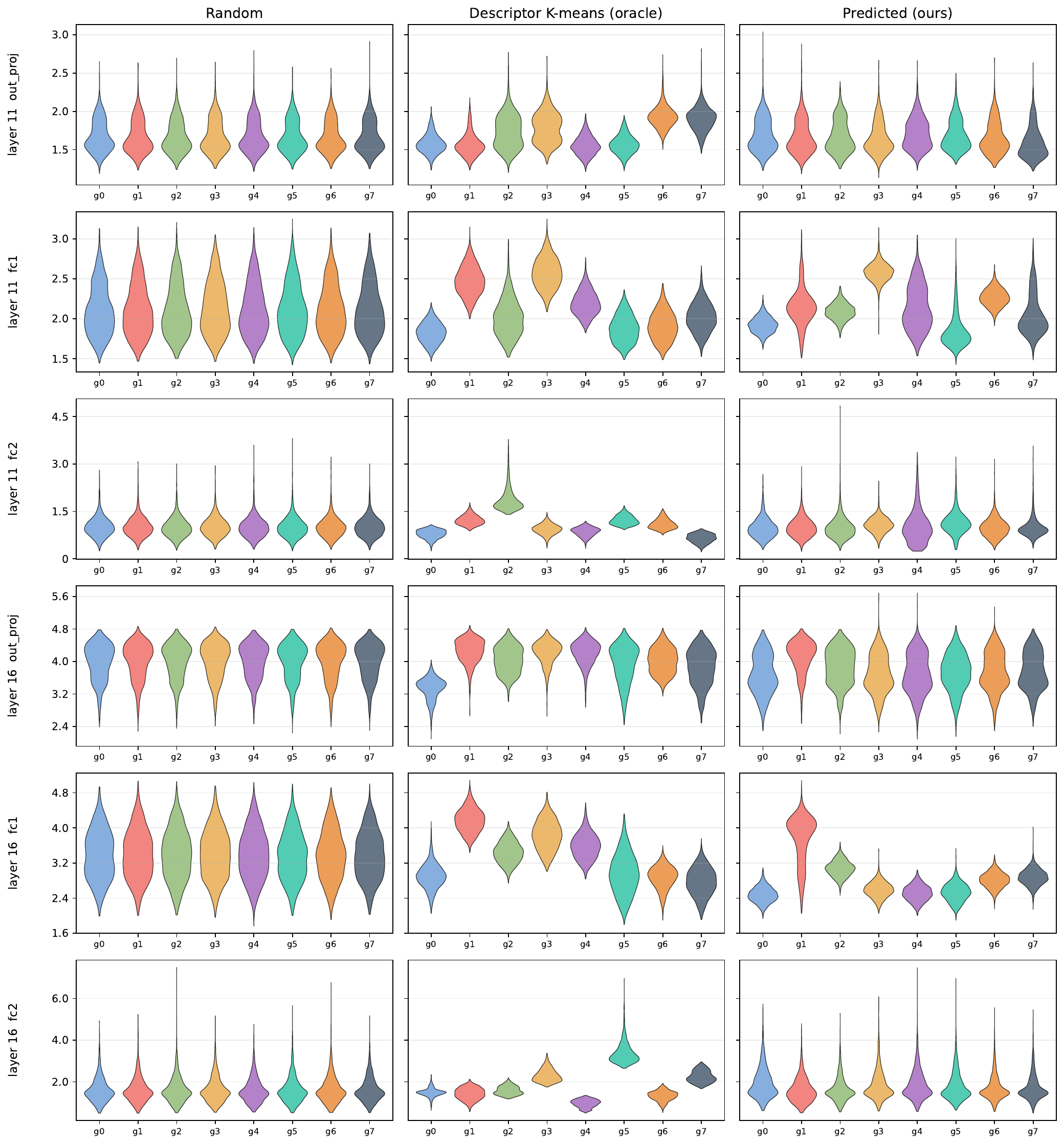}
  \caption{
  \textbf{Clipping-scale regimes in later layers.}
  Token-level preferred clipping scales $c_i^\star$ for layers 11 and 16 under
  random, post-scan descriptor K-means (oracle), and predicted groups.
  }
  \label{fig:app_clip_regime_l11_l16}
\end{figure*}

\clearpage


\newpage
\section*{NeurIPS Paper Checklist}



\begin{enumerate}

\item {\bf Claims}
    \item[] Question: Do the main claims made in the abstract and introduction accurately reflect the paper's contributions and scope?
    \item[] Answer: \answerYes{} 
    \item[] Justification: The abstract and introduction summarize the main contributions and scope, which are supported by our analysis, ablations, experiments, and latency results.

\item {\bf Limitations}
    \item[] Question: Does the paper discuss the limitations of the work performed by the authors?
    \item[] Answer: \answerYes{}.
    \item[] Justification: We discuss limitations and future work in the conclusion section.

\item {\bf Theory assumptions and proofs}
    \item[] Question: For each theoretical result, does the paper provide the full set of assumptions and a complete (and correct) proof?
    \item[] Answer: \answerNA{}.
    \item[] Justification: The paper does not present formal theoretical results.

    \item {\bf Experimental result reproducibility}
    \item[] Question: Does the paper fully disclose all the information needed to reproduce the main experimental results of the paper to the extent that it affects the main claims and/or conclusions of the paper (regardless of whether the code and data are provided or not)?
    \item[] Answer: \answerYes{}.
    \item[] Justification: We provide dataset, calibration, evaluation, quantization, implementation, and hardware details in the main paper and appendix.

\item {\bf Open access to data and code}
    \item[] Question: Does the paper provide open access to the data and code, with sufficient instructions to faithfully reproduce the main experimental results, as described in supplemental material?
    \item[] Answer: \answerYes{}.
    \item[] Justification: We use public benchmarks and provide anonymized source code in the supplementary material.

\item {\bf Experimental setting/details}
    \item[] Question: Does the paper specify all the training and test details (e.g., data splits, hyperparameters, how they were chosen, type of optimizer) necessary to understand the results?
    \item[] Answer: \answerYes{}.
    \item[] Justification: Experimental settings, hyperparameters, calibration details, and predictor training details are provided in the main paper and appendix.

\item {\bf Experiment statistical significance}
    \item[] Question: Does the paper report error bars suitably and correctly defined or other appropriate information about the statistical significance of the experiments?
    \item[] Answer: \answerNo{}.
    \item[] Justification: We do not report error bars, but evaluate across multiple datasets, scales, bit-widths, backbones, and restoration tasks.

\item {\bf Experiments compute resources}
    \item[] Question: For each experiment, does the paper provide sufficient information on the computer resources (type of compute workers, memory, time of execution) needed to reproduce the experiments?
    \item[] Answer: \answerYes{}.
    \item[] Justification: We report the GPUs, calibration time, iteration counts, trainable parameters, and latency on A6000 and Jetson Orin Nano.
    
\item {\bf Code of ethics}
    \item[] Question: Does the research conducted in the paper conform, in every respect, with the NeurIPS Code of Ethics \url{https://neurips.cc/public/EthicsGuidelines}?
    \item[] Answer: \answerYes{}.
    \item[] Justification: The research uses public image restoration benchmarks and does not involve human subjects, private data, or deceptive data collection.

\item {\bf Broader impacts}
    \item[] Question: Does the paper discuss both potential positive societal impacts and negative societal impacts of the work performed?
    \item[] Answer: \answerYes{}.
    \item[] Justification: The method can reduce edge deployment cost, while efficient image restoration may also be misused in surveillance or misleading image enhancement.
    
\item {\bf Safeguards}
    \item[] Question: Does the paper describe safeguards that have been put in place for responsible release of data or models that have a high risk for misuse (e.g., pre-trained language models, image generators, or scraped datasets)?
    \item[] Answer: \answerNA{}.
    \item[] Justification: The paper does not release high-risk models, scraped datasets, or systems with direct high-risk misuse potential.

\item {\bf Licenses for existing assets}
    \item[] Question: Are the creators or original owners of assets (e.g., code, data, models), used in the paper, properly credited and are the license and terms of use explicitly mentioned and properly respected?
    \item[] Answer: \answerYes{}.
    \item[] Justification: We use publicly available SR benchmarks (Set5, Set14, BSD100, Urban100, Manga109, DIV2K, DF2K) and pretrained MambaIRv1/v2 backbones, all under standard research-use licenses, and cite the original papers in the References. Baseline implementations follow the licenses of their public releases.


\item {\bf New assets}
    \item[] Question: Are new assets introduced in the paper well documented and is the documentation provided alongside the assets?
    \item[] Answer: \answerYes{}.
    \item[] Justification: We provide anonymized source code in the supplementary material.

\item {\bf Crowdsourcing and research with human subjects}
    \item[] Question: For crowdsourcing experiments and research with human subjects, does the paper include the full text of instructions given to participants and screenshots, if applicable, as well as details about compensation (if any)? 
    \item[] Answer: \answerNA{}.
    \item[] Justification: The paper does not involve crowdsourcing or human subjects.

\item {\bf Institutional review board (IRB) approvals or equivalent for research with human subjects}
    \item[] Question: Does the paper describe potential risks incurred by study participants, whether such risks were disclosed to the subjects, and whether Institutional Review Board (IRB) approvals (or an equivalent approval/review based on the requirements of your country or institution) were obtained?
    \item[] Answer: \answerNA{}.
    \item[] Justification: The paper does not involve human subjects.

\item {\bf Declaration of LLM usage}
    \item[] Question: Does the paper describe the usage of LLMs if it is an important, original, or non-standard component of the core methods in this research? Note that if the LLM is used only for writing, editing, or formatting purposes and does \emph{not} impact the core methodology, scientific rigor, or originality of the research, declaration is not required.
    \item[] Answer: \answerNA{}.
    \item[] Justification: LLMs are not used as an important or non-standard component of the core research method.

\end{enumerate}

\end{document}